\documentclass{article} 
\usepackage[preprint]{neurips_2024}
\usepackage{multirow}
\usepackage{tabularx}
\usepackage{booktabs}
\usepackage{makecell}
\usepackage{wrapfig}
\usepackage{textcomp}
\usepackage{tikz}
\usepackage{colortbl}
\usepackage{color}
\usepackage{xcolor}
\usepackage{subfigure}
\usepackage{pgfplots}
\usepackage{threeparttable}

\usepackage{amsmath,amsfonts,bm}

\def\eqref#1{equation~\ref{#1}}

\def\1{\bm{1}}

\DeclareMathAlphabet{\mathsfit}{\encodingdefault}{\sfdefault}{m}{sl}
\SetMathAlphabet{\mathsfit}{bold}{\encodingdefault}{\sfdefault}{bx}{n}

\usepackage{hyperref}
\usepackage{url}

\usepackage{amsmath}
\usepackage{amssymb}
\usepackage{pifont}
\newcommand{\cmark}{\ding{51}}%
\newcommand{\xmark}{\ding{55}}%

\usepackage{mathtools}
\usepackage{amsthm}
\theoremstyle{plain}
\newtheorem{takeaway}{Take-away}
\newtheorem{level}{Level}

\definecolor{petal1}{RGB}{255,0,0}    
\definecolor{petal2}{RGB}{0,255,0}    
\definecolor{petal3}{RGB}{0,0,255}    
\definecolor{petal4}{RGB}{255,255,0}  
\definecolor{petal5}{RGB}{255,0,255}  
\definecolor{dred}{RGB}{255, 33, 19}
\definecolor{dyellow}{RGB}{255, 211, 83}
\title{Roadmap towards Superhuman Speech Understanding 
using Large Language Models}

\author{Fan Bu$^1$$^\dagger$, Yuhao Zhang$^1$$^\dagger$, Xidong Wang$^1$, Benyou Wang$^1$\thanks{Benyou is the corresponding author (\textit{wangbenyou@cuhk.edu.cn}); $^\dagger$ means contributing equally.}~, Qun Liu$^2$, Haizhou Li$^1$\\
$^1$ The Chinese University of Hong Kong, Shenzhen\\
$^2$ Noah's Ark Lab, Huawei\\
\textit{wangbenyou@cuhk.edu.cn} \\
}
\begin{document}

\maketitle

\begin{abstract}

The success of large language models (LLMs) has prompted efforts to integrate speech and audio data, aiming to create general foundation models capable of processing both textual and non-textual inputs. Recent advances, such as GPT-4o, highlight the potential for end-to-end speech LLMs, which preserves non-semantic information and  world knowledge for deeper speech understanding.
To guide the development of speech LLMs, we propose a five-level roadmap, ranging from basic automatic speech recognition (ASR) to advanced superhuman models capable of integrating non-semantic information with abstract acoustic knowledge for complex tasks. 
Moreover, we design a  benchmark, \textbf{SAGI Bechmark}, that standardizes critical aspects across various tasks in these five levels, uncovering challenges in using abstract acoustic knowledge and completeness of capability. 
Our findings reveal gaps in handling paralinguistic cues and abstract acoustic knowledge, and we offer future directions. This paper outlines a roadmap for advancing speech LLMs, introduces a benchmark for evaluation, and provides key insights into their current limitations and potential.

\end{abstract}

\section{Introduction}

Paradigms to process \textit{language} have been reshaped thanks to LLMs and its scaling law. Given the success of LLMs, one may expect to integrate extensive data in \textit{speech} and \textit{audio} modality into LLMs (similar to visual language models~\cite{liu2023llava,li2023blip}~\footnote{There exists lighweight  solutions for adapting language models to process data beyond text (e.g., visual or auditory), such as: 1) using a lightweight encoder and alignment process, and 2) discretizing data into tokens, which supports the autoregressive objectives of LLMs.}), resulting in a more general foundation model. 

Towards this path, the exploration on speech foundation models recently brings new research insights from the perspectives of multi-task and multi-lingual processing ~\citep{whisper, bapna2021slam, zhang2023google,barrault2023seamless,pratap2024scaling}. A remarkable event is the release of GPT-4o, which is notable for its ability in open-ended speech-to-speech dialogue. Its performance in speech understanding, speech synthesis, and system latency has reached new levels, leading to a wave of studies on speech LLMs. The next question is, \textit{where are we now and where should we go?} To answer this, we begin by introducing the potential of using LLMs to understand speech.

\textbf{Processing Speech using LLMs}
Compared to the traditional approach of feeding ASR-transcribed text into text-only language models, unified speech-language models process raw audio or speech directly in an end-to-end fashion.
The \textit{benefits} for using LLMs to process speech are mainly two-fold.
\textbf{I) Preservation of non-semantic  information}: Processing raw speech directly through language models allows for the preservation of non-semantic  information, such as emphasis, speaker identity, background sounds, emotions, and feelings, to the greatest extent possible.

 \textbf{II) World knowledge inherited in LLMs}: LLMs have superior language understanding capabilities compared to traditional models and store vast amounts of world knowledge. Therefore, starting with an LLM as the foundation for speech processing allows for the natural inheritance of this embedded knowledge, which might benefit at speech recognition level.

\textbf{Five-level Speech Understanding}~

\begin{figure}[t]
    \centering
    \vspace{-15pt}
    \includegraphics[scale=0.30]{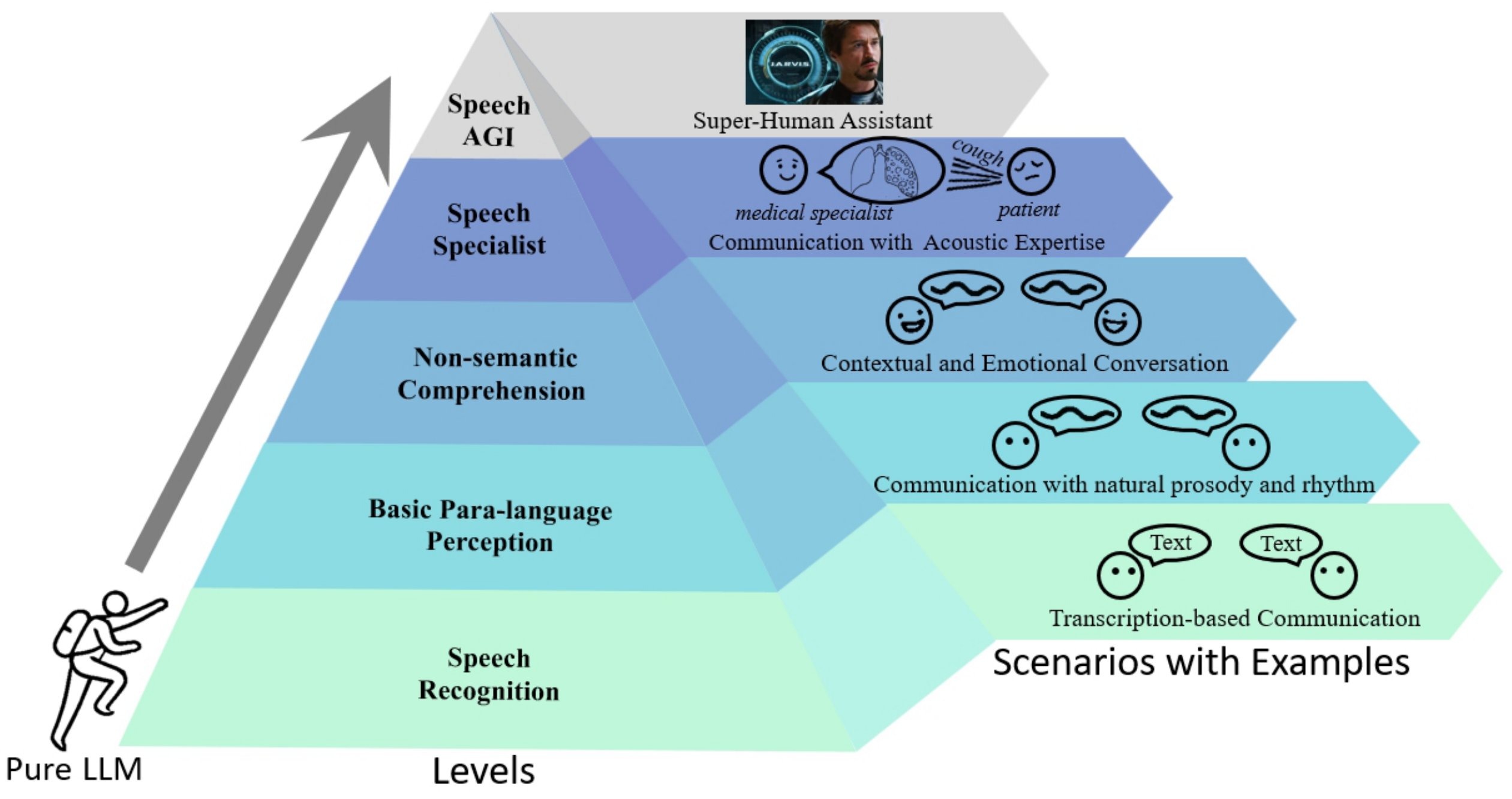}
    \caption{Levels of speech understanding using LLMs.}
    \label{fig:level}
\end{figure}

The two benefits highlight the potential of speech LLMs, achieving of which requires the models to perceive complete speech information and achieve abstraction  of  expert speech/acoustic knowledge (e.g.,  inferring from cough and melody in some applications).  To this regards, we define five levels (see Fig.~\ref{fig:level}.) as below:
\begin{itemize}
    \item \textbf{Basic Level}~ At the most basic level (\textbf{Level \ref{eq:level1}}), speech language models should  be able to  recognize speech as text. The rationale for defining automatic speech recognition as the foundational level is that it serves as the basis for directly interacting with LLMs through speech. However, these capabilities at the basic level (e.g., speech recognition) offer limited  additional benefits for ASR-equipped cascade model to understand human speech as it is somehow equivalent to a combination with a ASR model and a text-only LLM.
    \item \textbf{Acoustic Information Perception Levels}~ More advanced models (at \textbf{Level \ref{eq:level2}} and \textbf{Level \ref{eq:level3}}) are expected to directly perceive basic  paralinguistic information such as tone, pitch, and loudness, and further enable them to comprehend non-semantic cues like emotions and the surrounding environment (e.g., sarcasm).
    \item \textbf{Abstract Acoustic Knowledge Levels}~ At a higher level (at \textbf{Level \ref{eq:level4}}), models can integrate speech with expert  speech/audio knowledge to perform specialized tasks, such as medical assessments. At the final lavel (\textbf{Level \ref{eq:level5}}), the ultimate goal is to develop the \textbf{Speech Artificial General Intelligence (SAGI)} capable of combining non-semantic information with  speech/audio knowledge to perform all speech understanding tasks, even achieving superhuman speech understanding.
\end{itemize}

\textbf{The Benchmark} However, these levels remain insufficiently intuitive. Therefore, we have preliminarily developed a  benchmark to concretize and exemplify these capability levels. We designed the \textbf{SAGI Benchmark} to evaluate speech LLMs across various tasks that typically represent the characteristics of each level.

The benchmark covers a wide range of tasks, including speech recognition, language distinction, volume perception, emotion recognition, and more, with each task corresponds to a specific level of capability within speech LLMs. The reliability of these evaluation sets was verified using human test, open-source and custom-trained models, demonstrating that the tasks are feasible and can be accomplished. The benchmark aims to comprehensive, tiered evaluate speech LLMs' capabilities, and exploration of their ability to apply abstract acoustic knowledge.

\textbf{Findings}~
\textbf{Human} was generally strong in tasks from Level 1 to 3. However, at higher levels, human performance was limited due to a lack of abstract acoustic knowledge, which speech LLMs may start to outperform in certain tasks.

\textbf{The current speech LLMs}, though capable of surpassing human performance in a few areas, still fall short in terms of task diversity and comprehensiveness. Most models struggle with even basic paralinguistic information processing, highlighting the need for further improvement.
\textbf{We analyzed four reasons for performance deficiency} :  1) limited types of training data, 2) inability to comprehensively perceive acoustic information, 3) inadequate instruction following, and 4) weak LLM backbones.

The \textbf{contributions} of this paper are as follows: We propose a \textit{roadmap} to surpass human-level speech understanding, outlining five distinct levels to better characterize the current state of speech language models. Additionally, we design a \textit{benchmark} aligned with this roadmap, supplementing existing benchmarks with a variety of tasks. Finally, we present key \textit{findings} from the benchmark, based on evaluations of both speech LLMs and humans, and conduct a comprehensive \textit{analysis} of the factors behind their suboptimal performance, offering insights and guidance for future model and architecture development.

\section{Roadmap towards Understanding Speech}
\label{sec:roadmap}

To design a roadmap for future speech LLMs, we first analyzed the development process of speech LLMs in the past (in Sec.~\ref{sec:background}). Following that, we present our philosophy of the roadmap in Sec.~\ref{sec:philosophy_roadmap}.
\subsection{The Background}
\label{sec:background}
Current speech LLMs are mainly divided into two types: the Cascade Paradigm and the End-to-End Paradigm. Below, we will focus on analyzing these two approaches.

\begin{wrapfigure}[13]{rh}{0.475\textwidth}  
    \centering
    \vspace{-10pt}
    \resizebox{.41\textwidth}{!}{
    \begin{tikzpicture} 
    \node [anchor=center, align=center,rounded corners=2pt,fill=dyellow!30,minimum height=0.8cm,minimum width=1.8cm] (ASR) at (0,0) {ASR model};
    \node [anchor=south, align=center,rounded corners=2pt] (trans) at ([yshift=0.4cm]ASR.north) {Transcription};
    \node [anchor=center, align=center,rounded corners=2pt, rotate=90] (para) at ([yshift=1.1cm,xshift=-1.5cm]ASR.west) {Non-semantic \\ (emotion, stress, ... )};
    \node [anchor=south, align=center,rounded corners=2pt,fill=dred!30,minimum height=0.8cm,minimum width=1.8cm] (llms) at ([yshift=0.6cm]trans.north) {LLMs};
    \node [anchor=south, align=center,rounded corners=2pt] (output) at ([yshift=0.3cm]llms.north) {Output};
    \node [anchor=south, align=center,rounded corners=2pt] (times) at ([xshift=1.1cm,yshift=-1.55cm]para.east) {\textbf{\Large{$\times$}}};
    \node [anchor=north, align=center,rounded corners=2pt] (speech) at ([yshift=-0.3cm]ASR.south) {Speech input};
    
    \node [anchor=north, align=center,rounded corners=2pt] (cascade) at ([yshift=-0.3cm]speech.south) {\textbf{\textcolor{dred!80}{Cas}\textcolor{dyellow}{cade} method} };
    
    \draw[->,thick](speech.north)--(ASR.south);
    \draw[->,thick](ASR.north)--(trans.south);
    \draw[->,thick](trans.north)--(llms.south);
    \draw[->,thick](trans.north)--(llms.south);
    \draw[->,thick,dashed](ASR.north)..controls([yshift=0.2cm,xshift=-2.0cm]ASR.north) and ([yshift=0.8cm,xshift=1.2cm]para.east) ..(para.east);
    \draw[->,thick](llms.north)--(output.south);

    \node [anchor=west, align=center,rounded corners=2pt,fill=gray!30,minimum height=3.4cm,minimum width=2.4cm] (allms) at ([xshift=0.8cm,yshift=0.1cm]trans.east) {};
    \node [anchor=south, align=center,rounded corners=2pt,fill=dyellow!30,minimum height=0.8cm,minimum width=2.0cm] (ae) at ([yshift=0.2cm]allms.south) {Acoustic\\encoder};
    \node [anchor=south, align=center,rounded corners=2pt,fill=dred!30,minimum height=0.8cm,minimum width=2.0cm] (sllms) at ([yshift=0.7cm]ae.north) {LLMs};
    \node [anchor=north, align=center,rounded corners=2pt] (speech2) at ([yshift=-0.4cm]ae.south) {Speech input};
    \node [anchor=south, align=center,rounded corners=2pt] (output2) at ([yshift=0.95cm]sllms.north) {Output};
    \node [anchor=north, align=center,rounded corners=2pt] (all) at ([xshift=-0.65cm,yshift=0.05cm]allms.north) {Speech\\LLMs};
    \node [anchor=north, align=center,rounded corners=2pt] (cascade) at ([yshift=-0.3cm]speech2.south) {\textbf{\textcolor{gray!80}{End-to-end} method}};
    \draw[->,thick](speech2.north)--(ae.south);
    \draw[->,thick](ae.north)--(sllms.south);
    \draw[->,thick](sllms.north)--(output2.south);
    \end{tikzpicture}}

    \caption{\textit{Cascade}  and \textit{End-to-end} paradigms.}
    \label{Speechllmstypes}
\end{wrapfigure}
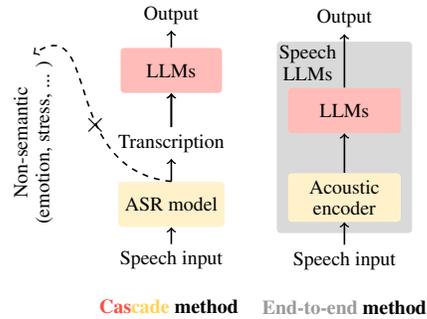

\textbf{Cascade Paradigm}~
A straightforward approach to understanding speech using LLMs is to feed speech transcriptions (in text format) into LLMs. This is known as the \textit{cascade} paradigm (see the left in Fig.~\ref{Speechllmstypes}). While this method allows for basic speech understanding, it lacks the ability to perceive non-semantic information (e.g., emotion, stress) within LLMs. This hinders a deeper understanding of the spoken content as its non-semantic information is often crucial for fully grasping the intent or nuances in speech.

\textbf{End-to-end Paradigm}~
In contrast, an \textit{end-to-end} speech LLM can process both semantic and non-semantic information simultaneously within a single model. This approach not only retains more detailed information within the LLM but also allows the world knowledge embedded in the LLM to interact directly with speech data. Note that this end-to-end speech paradigm introduces additional complexity, as it requires LLMs to handle raw speech data, which operates at a lower level compared to textual inputs.

In summary, the end-to-end solution enables LLMs to directly handle non-semantic information, such as emotions. Additionally, due to its stronger perceptual capabilities, it holds greater potential for understanding and applying abstract acoustic knowledge. As a result, end-to-end solution can be considered the future direction for the development of speech LLMs.

\begin{table}[t]
\small
\centering
\caption{Levels of speech understanding using LLMs}
\setlength{\tabcolsep}{0.6mm}{
    \resizebox{\textwidth}{!}
    {
\begin{tabular}{p{0.5cm}lcccp{4.5cm}}
\toprule

\multicolumn{2}{l}{  \multirow{2}{*}{\textbf{Level}}}
 & \textbf{Semantic} & \textbf{Non-Semantic  } &\textbf{Abstract Acoustic} &   \makecell[c]{\multirow{2}{*}{\textbf{Remark}}} \\
 & & \textbf{Information}  & \textbf{Information} & \textbf{Knowledge}   \\
\midrule
- &   \textbf{Pure LLM}  & - & - & - & Without speech input. \\
\midrule
L1 & \textbf{Basic ASR  } &  \cmark  & \xmark & \xmark & Recognizing Speech as texts. \\
\midrule
L2 & \makecell[l]{\textbf{Paralinguistic}\\ \textbf{Perception }} &  \cmark &  only paralinguistic & \xmark &  \makecell[l]{Perceiving direct paralinguistic \\ \textit{information} like tone, pitch,\\ loudness, rhythm, and speech rate.}\\
\midrule
L3 & \makecell[l]{\textbf{Non-semantic} \\ \textbf{Comprehension }}  &  \cmark & \cmark &   \xmark & \makecell[l]{ Comprehending non-semantic \\ \textit{information} like speaker identity, \\ gender, age, emotional state, and \\environmental sounds.} \\
\midrule
L4 & \makecell[l]{\textbf{Speech}\\ \textbf{Specialist} }&   \cmark & \cmark &   specialist & \makecell[l]{Understanding speech with \textit{specific} \\speech \textit{knowledge}.}\\
\midrule
L5 &\makecell[l]{\textbf{Speech  AGI}\\ \textbf{(Generalist) } }&     \cmark & \cmark &  generalist & \makecell[l]{Understanding speech with \textit{general}\\ speech \textit{knowledge}.}\\
\bottomrule
\end{tabular}
}
}
\label{tab:speech_understanding_levels}
\end{table}

\subsection{The Philosophy of the Roadmap}
\label{sec:philosophy_roadmap}

With the rise of large language models (LLMs), there is an increasing demand to understand information beyond text, particularly speech. The core idea is that speech conveys richer information than text alone, positioning ASR (Automatic Speech Recognition) as a foundational level. End-to-end speech LLMs can begin with ASR capabilities to directly leverage the capabilities of text LLMs. And then, it progressively incorporate more advanced comprehension of non-semantic features. Finally it contains the ability to retain and apply abstract acoustic knowledge. This progress can be described as evolving through the following five levels:

\begin{level}
\label{eq:level1}
\textbf{Speech Recognition Level}~
  At the most basic level, a speech language model should be capable of recognizing text.
\end{level}

These tasks form the most fundamental requirements for interacting with large models using speech. However, even  at Level \ref{eq:level1}, the model offers limited advantages over a traditional cascade approach (e.g., feeding ASR-transcribed text into LLMs). The real benefits of speech LLMs begin to emerge at the next level, with the ability to capture non-semantic features such as paralinguistic information.

\begin{level}
\label{eq:level2}
\textbf{Basic Paralinguistic Perception Level}~
At this level, Speech LLMs gain the ability to perceive basic paralinguistic features in speech, such as tone, pitch, volume, rhythm, and speech rate. 
\end{level}

These elements are essential to speech comprehension and provide distinct advantages over pure text-based models (or Speech LLMs at Level \ref{eq:level1}). While this lays the foundation for more advanced capabilities, the insights derived at this level are still relatively shallow. For deeper understanding, we must move to Level \ref{eq:level3}, where a model comprehends a broader range of non-semantic information.

\begin{level}
\label{eq:level3}
\textbf{Non-semantic Comprehension Level}~
  At this stage, the Speech LLM extends beyond basic paralinguistic features and is capable of  comprehending and interpreting more complex non-semantic information, such as emotions, sarcasm, and heightened states like pride. 
\end{level}

For example, 
emotions are higher-level human experiences that involve cognitive functions, distinguishing them from basic paralinguistic information. Interestingly, even some higher animals, like pet dogs, can perceive these types of non-semantic information. To fundamentally distinguish humans from  animals, we designed Level~\ref{eq:level4}~ by leveraging the human strengths in higher-level cognitive capabilities.

\begin{level}
\label{eq:level4}
\textbf{Speech Specialist Level}~
  At this advanced level, Speech LLMs integrate expert-level abstract acoustic knowledge to handle a few specific, complex tasks.
\end{level}

This requires integrating abstract acoustic knowledge which are advanced knowledge derived from acoustic information. 
This goes beyond mere recognition and comprehension at Level \ref{eq:level1} and Level \ref{eq:level2}, requiring the model to apply higher-order thinking skills (such as analysis, evaluation, and creation) based on acoustic information~\footnote{This capability benefits a range of tasks, such as: 1) using cough sounds to identify the type and origin of the cough, 2) pronunciation correction, 3) music appreciation, 4) stethoscope auscultation, 5) early screening for depression and Parkinson's disease, and 6) understanding animal vocalizations.}, according to Bloom’s cognitive taxonomy~\cite{krathwohl2002revision}.
Despite these abilities, the model at this level remains domain-specific, which leads to the need for a fully generalized Speech LLM, as defined by Level \ref{eq:level5}.

\begin{level}
\label{eq:level5}
\textbf{Speech AGI level}~
The ultimate level, Speech Artificial General Intelligence (SAGI), represents a comprehensive speech model that functions as a generalist. It can integrate knowledge from various domains and perform both general and specialized tasks, potentially surpassing human experts.  
\end{level}
This vision of SAGI represents the culmination of speech understanding, combining domain expertise, adaptability, and the capacity to exceed human performance in speech-based tasks.
SAGI’s potential to outperform humans probably stems from its ability to scale  learning time and superior memory retention compared to humans. Due to time constraints, humans can typically only specialize in a narrow domain, as illustrated by `\textit{The 10,000-Hour Rule}' in Malcolm Gladwell's book Outliers. In contrast, LLMs can easily scale their learning time by leveraging larger computing resources. Furthermore, LLMs generally possess longer memory—whether explicit or implicit—than humans, enhancing their ability to retain and apply vast amounts of information.

\section{Benchmarking}

\label{headings}

\subsection{The New Benchmark: SAGI}

To implement the roadmap (Sec.\ref{sec:roadmap}), we aim to build a comprehensive benchmark to concretes these levels.
Though previous benchmarks for speech LLMs have contributed significantly, they focus mainly on the first three levels, neglecting abstract acoustic knowledge and broader SAGI applications (App.\ref{sec:existing}). Additionally, current benchmarks lack the depth needed for full speech LLM development, particularly in foundational tasks like pitch and volume perception. To address these gaps, we propose a new benchmark, detailed in the following section.

\begin{table}[h]
\centering
\scriptsize
\caption{Overview of the levels and the corresponding tasks.}
\label{tab:dataset_detail}
\begin{tabular}{@{}clp{8.5cm}}
\toprule
\textbf{Level} & \textbf{Task} & \textbf{Dataset} \\ \midrule
\multirow{5}{*}{\textbf{L1}} & Language Identification & Europarl-ST \citep{Europarl-ST} \\
                     & Auto-Speech Recognition & LibriSpeech \citep{LibriSpeech} \\
                     & ASR for Legal Terms$^*$ &  Made of CosyVoice \citep{funaudiollm}\\
                     & ASR for Medical Terms$^*$&  Made of CosyVoice \citep{funaudiollm}\\
                     & Auto-Lyrics Transcription & Jam-Lyrics \citep{Jam-Lyrics} \\ \midrule
\multirow{3}{*}{\textbf{L2}} & Volume Perception & Made of LJSpeech \citep{ljspeech} \\
                     & Pitch Perception & Made of SpeechAccentArchive \citep{SpeechAccentArchive}\\
                     & Binaural Effect Perception & Our proposed method \\ \midrule
\multirow{9}{*}{\textbf{L3}} & Ambient Sound Detection & Noisy speech \citep{noisySpeech} \\
                     & Acoustic Scene Classification & Made of MS-SNSD \citep{MSSNSD} \\
                     & Speaker’s Age Prediction & Made of AIR-Bench \citep{airbench} \& SpeechAccentArchive \citep{SpeechAccentArchive} \\
                     & Speaker’s Gender Recognition & VCTK \citep{vctk} \\
                     & Speech Emotion Recognition & Selected from RAVDESS \citep{ravdess} \\
                     & Cappella Emotion Recognition & Selected from RAVDESS \citep{ravdess} \\
                     & Emotional Intensity Perception & Made of RAVDESS \citep{ravdess}  \\ 
                     
                    & Emotion Translation$^*$ & Made of RAVDESS \citep{ravdess} and CosyVoice \citep{funaudiollm}\\
                      & Singing Detection & RAVDESS \citep{ravdess} \\
                     \midrule
                     
\multirow{4}{*}{\textbf{L4}} 
                     &COVID-19 Risk Detection  &Virufy  \citep{virufy} \\
                     & Cough Type Classification & Made of COUGHVID\citep{coughvid}\\
                     & Cough Origin Diagnosis & Made of COUGHVID\citep{coughvid}\\
                     & Cough Severity Assessment & Made of COUGHVID\citep{coughvid}\\
                     \midrule
\multirow{2}{*}{\textbf{L5}} & Spoken English Coach & Made of speechocean762 \citep{speechocean762} \\
                     & Voice Detective & Made of SpeechAccentArchive \citep{SpeechAccentArchive} \\
\bottomrule
\end{tabular}

``*'' denotes that utterances are synthesized, and the credibility verification is provided in Appendix~\ref{app:cred_verify}.
\end{table}

\textbf{Philosophy of Benchmark}~
The SAGI Benchmark is structured to align with the five levels of speech understanding\footnote{The types of tasks for Level \ref{eq:level4} and \ref{eq:level5} are not yet complete in the current version; we are working on adding more diverse tasks.}, and the overview of the benchmark is shown in Tab.~\ref{tab:dataset_detail}. 
The tasks are organized into five levels: \textbf{Level 1} focuses on testing the \textbf{recognition capabilities} of speech LLMs, including ASR, lyrics transcription, and term recognition tasks. \textbf{Level 2} evaluates \textbf{foundational perception} abilities, such as pitch and volume perception for tasks like age, gender, and emotion recognition. \textbf{Level 3} assesses \textbf{non-semantic comprehension}, incorporating tasks like emotion-integrated translation, environment perception, and emotional intensity recognition. \textbf{Level 4} explores the application of \textbf{abstract acoustic knowledge}, specifically focusing on medical-related contexts. Finally, \textbf{Level 5} envisions the capabilities of \textbf{Speech AGI (SAGI)}, highlighting tasks that promote creativity and diverse thinking, such as appreciating artwork, with a strong foundation in earlier levels.

\subsection{Benchmarked Objects}

\label{eval_process}
\textbf{Humans}~
To conduct an initial evaluation of human performance, we created evaluation subsets by randomly selecting 80 samples per label for the objective multiple-choice tasks, and 80 samples in total for the other tasks. Four students (two males and two females) with strong English proficiency completed the assessments. The results are recorded in Tab.~\ref{main_results}. The participant information and consistency test is in App.~\ref{sec:humans_detail}.

\textbf{Speech LLMs}~
There are four types of speech LLMs, see more details in  Sec.~\ref{gen_inst}. We selected an open-source model for each type, except for video LLMs, where the performance on audio-only tasks is not stable. For speech-related models, we chose Qwen2-Audio for its strong performance. We selected Mu-llama for the music model and GAMA for the audio model. Additionally, we tested SALMONN as a mixed audio and speech model. We further test GPT-4o advanced speech mode. Because only some models supports the speech instruction, we utilize the text instruction to ensure fair comparison.

For more details on model replication and evaluation settings, please refer to App.~\ref{appendix:model-evluation}.

\subsection{Benchmarking Results}
\label{eval_result}

\begin{table}[th]
    \centering
    \scriptsize
    \caption{Performance of Speech LLMs on SAGI Benchmark. }

    \label{main_results}
    \resizebox{\textwidth}{!}
    {
    \begin{threeparttable}

    \begin{tabular}{@{}clcccccc@{}} 
        \toprule
        \multirow{2.5}{*}{\makecell[l]{\textbf{Level}}}&\multirow{2.5}{*}{\textbf{Task}} & \multirow{2.5}{*}{\makecell{\textbf{Human
        }\\ \textbf{Baseline}}} &\multicolumn{5}{c}{\textbf{Models}} \\
        \cmidrule(l){4-8}
 & & & GPT-4o & MuLLaMA & GAMA & SALMONN & Qwen2-Audio \\
\midrule
\multirow{6}{*}{\textbf{L1}}&Language Identification & \texttimes & 88.50\% &\ \  8.48\% & \texttimes & 35.17\% & 96.44\% \\
&Auto-Speech Recognition & 15.49$^*$ & 10.24$^*$  & \texttimes & \texttimes & 5.45$^*$ & 4.63$^*$ \\
&ASR for Legal Terms & 98.50\% & \ \ 26.47\%& \texttimes & \texttimes & \texttimes & 81.04\% \\
&ASR for Medical Terms & 97.50\% & 41.87\% & \texttimes & \texttimes & \texttimes & 53.86\% \\

&Auto-Lyrics Transcription  & 26.88$^*$ & \texttimes \  & \texttimes & \texttimes & 77.12$^*$ \ \  & 32.48$^*$ \ \  \\
& \ \ - Hallucination Rate &\ \  3.00\% & \texttimes  & \texttimes & \texttimes & 29.26\%  & 38.21\%  \\
\midrule
\multirow{3}{*}{\textbf{L2}}&Volume Perception & 100.00\% &\texttimes & 50.00\% & 11.98\% & 53.22\% & 48.96\% \\
&Pitch Perception & \ \ 96.25\% & 29.33\%& 33.78\% & 41.50\% & 50.00\% & 50.00\% \\
&Binaural Effect Perception & 100.00\% & 41.38\% & \texttimes & \texttimes & 49.88\% & \texttimes \\
\midrule
\multirow{9}{*}{\textbf{L3}}&Ambient Noise Detection & 91.88\% & 45.27\% & 50.00\% & 60.17\% & 49.88\% & 50.00\% \\
&Acoustic Scene Classification & 90.28\% & 16.36\% & \ \ 5.07\% & 12.05\% & 20.74\% & 27.67\% \\

&Speaker’s Age Prediction& 52.59\% & 13.43\% & 33.60\% & \texttimes & 36.87\% & 38.55\% \\
&Speaker’s Gender Recognition& 97.50\% & \texttimes & 50.00\% & \texttimes & 48.12\% & 79.60\% \\

&Speech Emotion Recognition & 50.71\% & 16.77\% & \ \ 9.20\% & \ \ 3.68\% & 10.93\% & 79.51\% \\
&Cappella Emotion Recognition & 62.25\% & 21.50\% & 12.42\% & \ \ 7.08\% & 14.62\% & 62.38\% \\
&Emotion Intensity Perception & 97.50\% & 72.67\% & 50.00\% & 50.00\% & 49.29\% & 50.00\% \\
&Emotion Translation$^\dagger$ & 3.68 & 0.32\ \   & \texttimes & \texttimes & 0.27\ \  & 0.31\ \  \\
&Singing Detection & 99.38\% & 53.11\% & 50.00\% & 64.82\% & 56.47\% & 50.22\% \\

\midrule
\multirow{4}{*}{\textbf{L4}}

&COVID-19 Risk Detection & 60.63\% & \texttimes & \texttimes & \texttimes &  50.00\% & 14.17\% \\
&Cough Type Classification & 52.50\% & 40.33\% & 50.16\% & 44.17\% & 49.17\% & 43.39\% \\
&Cough Origin Diagnosis & 32.19\% & \texttimes & \texttimes & \texttimes & \ \ 4.01\% & 25.65\% \\
&Cough Severity Assessment & 45.42\% & 24.12\% & 30.85\% & 28.50\% & 38.24\% & 33.86\% \\
\midrule
\multirow{2}{*}{\textbf{L5}}&Spoken English Coach$^\dagger$ & 1.39 & 0.15  & 1.29\ \  & 0.44\ \  & 0.48\ \  & 0.54\ \  \\
&Voice Detective$^\dagger$ & 1.20 & \texttimes  & 0.84\ \  & 0.83\ \  & 0.86\ \  & 1.24\ \  \\
        \bottomrule
    \end{tabular}
    
\begin{tablenotes}[para,flushleft]
``$\times$'' indicates that the model fails to follow the instruction. ``*'' denotes that the metric is Word Error Rate (WER) and similar metrics, for which lower values indicate better performance. ``$\dagger$'' indicates that the task is evaluated by GPT-4, with a score ranging from 1 to 4. 

\end{tablenotes}
\end{threeparttable}
    }
\end{table}

\textbf{Humans}
\label{section:human_perfomance}
~As seen in Tab. \ref{main_results}, human performs generally well from Level \ref{eq:level1} to \ref{eq:level3}. However, it becomes worse at higher levels due to a lack of acoustic knowledge. On the other side, speech understanding for humans are generally better than speech language models.

\begin{takeaway}
\textbf{Human performance:}~
\textit{Human generally performs well in speech understanding from Level 1 to 3, but fails to reach a high level due to a lack of abstract acoustic knowledge.} 
\end{takeaway}

\textbf{Speech LLMs}
\label{sec:SLLM performance}
~As shown in Tab.~\ref{main_results}, speech LLMs exhibit a significant weakness in Level \ref{eq:level2} which consists of basic listening abilities of the human. These models are currently focused on directly addressing high-level tasks while neglecting basic paralinguistic information perception, thereby the model fails to shows generalization at higher level. Furthermore, most models do not fully satisfy the requirements at any given level, highlighting a lack of consideration for both task diversity and comprehensiveness. 
Notably, Qwen2-Audio has outperformed humans in tasks like emotion recognition. This suggests that speech LLMs have the potential to detect subtle changes in speech, even beyond human capabilities.

\begin{takeaway}
\textbf{Speech LLMs:}~
Speech LLMs still struggle with non-semantic perception and comprehension from Level 1 to Level 3, despite excelling in some tasks, limiting their performance on more complex tasks at higher levels.
\end{takeaway}

\textbf{GPT-4o} 
~The results indicate that GPT-4o tends to reject audio-related tasks. Compared to other models, GPT-4o shows merit in emotion-related tasks but fails to demonstrate overwhelming advantages in understanding ability. We suppose its strength lies in its interaction capability. Therefore, we tested its ability to follow speech instructions, which directly evaluates its interaction skills. We also tested Qwen2-Audio, one of the few models that support speech instructions.

 \begin{table}[h]
     \centering
     \footnotesize
     \caption{Comparison of performance based on text instructions and speech instructions.}
     \begin{tabular}{@{}lcc|cc@{}} 
        \toprule
        \multirow{2.5}{*}{\textbf{Task}}& \multicolumn{2}{c}{\textbf{Text instructions}}& \multicolumn{2}{c}{\textbf{Speech instructions}}\\
        \cmidrule(l){2-5}
        &GPT-4o&Qwen2-Audio&GPT-4o&Qwen2-Audio\\
        \midrule
        Language Identification & 88.50\%& \textbf{93.01\%}& \ \ \textbf{91.45\%}&\ \ 18.64\%\\
        Auto-Speech Recognition  & 10.24\ \ \ \  &\textbf{4.63}\ \ &\textbf{14.65}\ \ \ &22.39\ \ \\
        Speech Emotion Recognition &16.77\% &\ \ \textbf{79.51\%}\ \ & \ \ \textbf{23.46\%}&\texttimes\\
        Emotion Intensity Perception& \ \textbf{72.67\%} &50.00\%&\ \ \textbf{10.84\%}&\texttimes \\
        \bottomrule
     \end{tabular}
     \label{tab:speechinstruction_textinstruction}
     \\We selected tasks in which at least one model performed well under text instruction conditions. \\Details about the speech instruction can be found in App.~\ref{app:speech-instruction}.
\end{table}

The performance is detailed in Tab.~\ref{tab:speechinstruction_textinstruction}. Compared to the results with text instructions, GPT-4o performs better with speech instructions, while Qwen2-Audio loses most of its capabilities. However, there remains a significant gap compared to the best results achieved using text instructions.

\begin{takeaway}
\textbf{GPT-4o:} ~GPT-4o demonstrates clear advantages in following speech instructions, but there is still significant room for improvement. 
\end{takeaway}

\textbf{Future Prospects}
\label{future}

We observe that abstract acoustic knowledge presents a common bottleneck for both humans and speech LLMs in reaching higher performance levels. Given superior capabilities of LLMs in knowledge acquisition, meanwhile, the deficiencies in diversity and completeness of capabilities can be ameliorated by incorporating additional training data. we contend:
\begin{takeaway}

\textit{\textbf{Speech LLMs have the potential to exceed human capabilities,} yet they currently fall short in addressing the full scope of tasks and integrating abstract acoustic knowledge. }
\end{takeaway}

\section{More Analysis on Performance Deficiency}
\label{sec:reasons}

In this section, we discuss reasons for performance deficiency in SAGI benchmark. We first consider composition of training data (in Sec.~\ref{sec:task}). Then we analyse the model from three perspectives: 1) perception of acoustic information (in Sec.~\ref{sec:information_transfer}), 2) ability of instruction following (in Sec.~\ref{sec:text_input}), and 3) capacity of LLM backbone (in Sec.~\ref{foundation LLMs}).

\subsection{Limited Types of Training Data }
\label{sec:task}

\begin{figure}[th]
    \centering
    \includegraphics[scale=0.30]{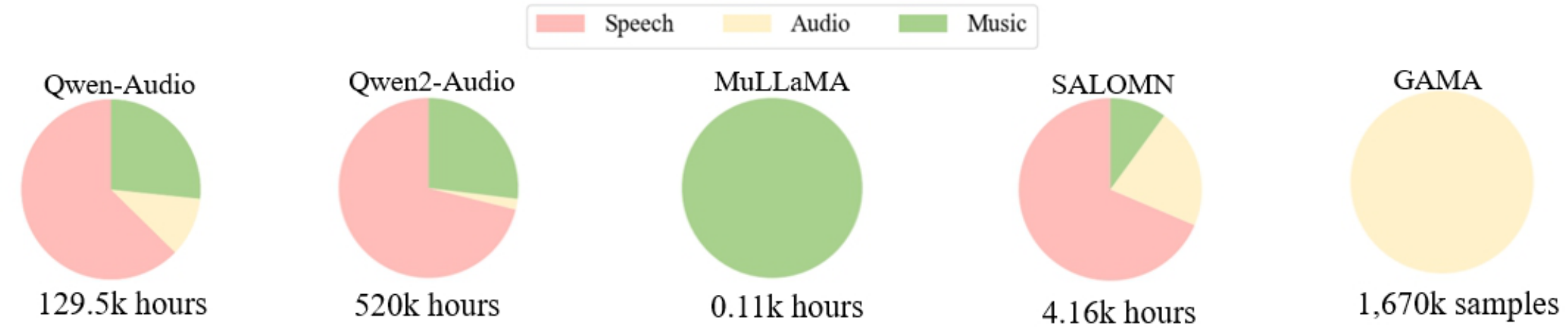}
    \caption{Distribution of three types of training data used by various models}
    \label{fig:training_data_sector}
\end{figure}

We observed in Tab.~\ref{main_results} that certain tasks, particularly those in Level \ref{eq:level2}, are easy for humans but challenging for speech LLMs. We first analyzed the composition of the training data for speech LLMs, as shown in Fig.~\ref{fig:training_data_sector}. We found that most speech LLMs tend to disregard audio data except for GAMA, whereas GAMA focuses primarily on audio. This indicates distinct data biases among different speech LLMs, leading to variations in task preferences.

To further examine the influence of task preference, we compared the performance of various speech LLMs with Whisper V3 (trained with $\sim$5,000k hours), as shown in Tab.~\ref{tab:baseline}. We found that Whisper still outperforms other models on the Lyrics Transcription task  due to its the massive training data. On the other hand, with the help of the learned knowledge, speech LLMs perform significantly better at recognizing certain terms. This demonstrates that speech LLMs have great potential compared to traditional speech models. Notably, we also tested a \textit{Small} model trained exclusively on an audio dataset. This \textit{Small} model achieved 100\% accuracy, while speech LLMs struggled with the task.

 \begin{table}[h]
     \centering
     \footnotesize
     \caption{Comparison of task-specific model and LLMs.}
     \begin{tabular}{@{}lllcc@{}} 
        \toprule
        \textbf{Task} &\textbf{Task type}& \textbf{Model}& \textbf{Result} & \textbf{Best result of LLMs}\\
        \midrule
        Language Identification & 5-Categories& Whisper& \ \ 91.45\%&\textbf{96.62\%}\\
        Auto-Speech Recognition  &Generation &Whisper&\textbf{2.44}&4.63\ \ \\
        Auto-Lyrics Transcription &Generation& Whisper & \textbf{22.10} \ \ &32.48\ \ \  \\
        ASR for Legal Term &Generation &Whisper&\ \ 33.33\%&\textbf{\ \ 81.04\%} \\
        ASR for Medical Term &Generation &Whisper&\ \ 34.98\%&\textbf{\ \ 53.86\%} \\
        
        \midrule
        Volume Perception & 2-Categories&Small model& \ \ \textbf{100.00\%}&\ \ 53.22\% \\
        \bottomrule
     \end{tabular}
     \label{tab:baseline}
     \\The \textit{Small} model uses Transformer with 10M parameters.
\end{table}

\begin{takeaway}
\textit{\textbf{Current insufficient diversity and completeness of training data could not help speech LLMs reach a higher level.}}
\end{takeaway}

\subsection{Inability to Comprehensively  Perceive Acoustic Information}

\label{sec:information_transfer}
The current end-to-end paradigm universally adopts the stacking paradigm. However, the stacking paradigm may suffer from two types of information loss: 1) the latent representation produced by the acoustic encoder does not fully capture or convey the necessary information, and 2) the acoustic encoder fails to transfer all the information to the downstream LLMs. 
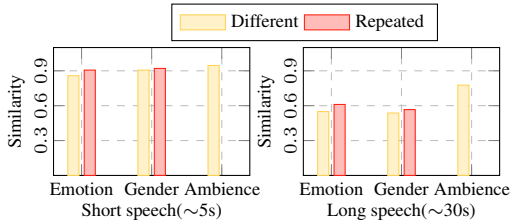
\begin{wrapfigure}[14]{r}{0.5\textwidth}
  \centering
  \begin{tikzpicture}
    \scriptsize{
    \begin{axis}[
      at={(0,0)},
      ymajorgrids,
      xmajorgrids,
      grid style=dashed,
      legend style={at={(0.6,1.05)}, anchor=south west},
      legend columns=-1,
      ybar,
      xtick align=inside,
      height=.235\textwidth,
      width=.3\textwidth,
      bar width=0.6em,
      xlabel={Short speech($\sim$5s)},
      ylabel={Similarity},
      symbolic x coords={{Emotion}, {Gender}, {Ambience},},
      xticklabel style={text width=1cm, align=center},
      xtick=data,
      nodes near coords align={vertical},
      ymin=0.,
      ymax=1.1,
      ytick={0.3,0.6,0.9},
      legend entries={Different, Repeated},
      enlarge x limits=0.2,
      ylabel style={yshift=-3em},xlabel style={yshift=1.0em},
      yticklabel style={/pgf/number format/fixed,/pgf/number format/fixed zerofill,/pgf/number format/precision=1,rotate=90},
      ]
          \addplot[fill=dyellow!30, draw=dyellow,area legend] coordinates {({Emotion},0.8579) ({Gender},0.9060) ({Ambience}, 0.9456)  };

          \addplot[fill=dred!30, draw=dred,area legend] coordinates {({Emotion},0.9060) ({Gender},0.9205)   };

    \end{axis}
    }

    \scriptsize{
    \begin{axis}[
      at={(13.5em,0)},
      ymajorgrids,
      xmajorgrids,
      grid style=dashed,
      legend style={at={(0.01,0.68)}, anchor=south west},
      ybar,
      xtick align=inside,
      height=.235\textwidth,
      width=.3\textwidth,
      bar width=0.6em,
      xlabel={Long speech($\sim$30s)},
      ylabel={Similarity},
      symbolic x coords={{Emotion}, {Gender}, {Ambience}},
      xtick=data,
      xticklabel style={text width=1cm, align=center},
      nodes near coords align={vertical},
      ymin=0.0,
      ymax=1.1,
      ytick={0.3,0.6,0.9},
      enlarge x limits=0.2,
      ylabel style={yshift=-3em},xlabel style={yshift=1.0em},
      yticklabel style={/pgf/number format/fixed,/pgf/number format/fixed zerofill,/pgf/number format/precision=1,rotate=90},
      ]
          \addplot[fill=dyellow!30, draw=dyellow,area legend] coordinates {({Emotion},0.5492) ({Gender},0.5369) ({Ambience},0.7767)  };

          \addplot[fill=dred!30, draw=dred,area legend] coordinates {({Emotion},0.6105) ({Gender},0.5662) };

    \end{axis}
    \vspace{-10pt}
    }

\end{tikzpicture}
\caption{Representation similarity of different speeches. Each speech pair has the same content but
is spoken in a different style. The representation is generated by the Whisper encoder.
}\label{similarity}
\end{wrapfigure}

We first investigate whether the loss of latent representation contributes to the limited performance. We compare the speech features generated from the same text content, which are spoken by different genders and with different emotions. The features are generated by Whisper, and cosine similarity is used to analyze the original and perturbed speech. The results, shown in Fig.~\ref{similarity}, indicate that there is no significant difference between different speech samples. This suggests that emotion and gender information is lost during the acoustic encoder process. This could explain why some speech LLMs perform poorly on certain simple tasks, as the LLMs cannot compensate for the loss of acoustic information.

We then assess whether information loss from the acoustic encoder to downstream LLMs limits speech LLMs' performance. We select cases from the ASR task where the WER is higher than 20\%, as shown in Tab.~\ref{tab:ASR_error}. We found that the error types is different between the whisper and speech LLMs. Considering that Qwen2-Audio is built on Whisper, the results confirm that LLMs cannot correct errors from the acoustic model. A notable difference between Whisper and speech LLMs is the tendency of the latter to produce overlong outputs, which is a form of hallucination.

\begin{table}[h]
\small
\vspace{-9pt}
     \centering
     \caption{Two types of recognizing errors. The ``truncation'' and ``over-long'' denote the generation is short and longer than the length of reference more than 20\% separately.}
     \begin{tabular}{@{}lrrr@{}} 
        \toprule
        \textbf{Model}&\textbf{Total}&\textbf{Truncation}&\textbf{Over-long}\\
        \midrule
Whisper&64&3&0\\
Qwen-Audio&68&5&6\\
Qwen2-Audio&149&89&3\\
SALMONN & 251&154&5\\
        \bottomrule
     \end{tabular}
     \label{tab:ASR_error}
     
\end{table}

Another notable phenomenon is that almost 60\% of error cases are caused by truncation. Additionally, we observed that the speech LLMs sometimes omits the start of a sentence, which does not happen with Whisper. This proves that speech LLMs suffer the loss of information transfer between the LLMs and the acoustic encoder. The current stacked paradigm often tunes base on LLMs with most parameters frozen, which requires the acoustic features to fit the LLMs' representation space. This requirement hinders the seamless transmission of acoustic information to the LLMs, leading to premature termination of the generation process.

\begin{takeaway}
\textit{\textbf{LLMs in current end-to-end solutions fail to encode complete acoustic information.}}
\end{takeaway}

\subsection{Inadequate Instruction Following }
\label{sec:text_input}
We observed that some models exhibit poor instruction following in Tab.~\ref{main_results}. Two reasons can lead to these results: 1) the models do not understand the instructions, and 2) the instruction fails to help the models comprehend the speech. 

\begin{wrapfigure}[15]{r}{0.50\textwidth}
	\centering
        \vspace{-8pt}
		\begin{tikzpicture}
		\scriptsize{
			\begin{axis}[
				at={(0,0)},
				ymajorgrids,
				xmajorgrids,
				grid style=dashed,				
				legend style={at={(-0.01,0.92)}, anchor=south west},
				legend columns=-1,
				xlabel={\scriptsize{\#Instruction}},
				xlabel style={yshift=1.2em},
				ylabel={\scriptsize{Accuracy/\%}},
				ylabel style={yshift=-1.5em},
				yticklabel style={/pgf/number format/precision=1,/pgf/number format/fixed zerofill},
				height=.26\textwidth,
				width=.29\textwidth,
				ymin=0.0,ymax=45, ytick={5,15,25,35},
				xmin=0,xmax=10,xtick={2,4,6,8},
				legend style={yshift=8pt,xshift=-4.4em, legend plot pos=right,cells={anchor=west}}
				]
				
			\addplot[orange!80,mark=triangle*,mark size=1.5pt,thick,mark options={fill=white,draw=orange,line width=0.5pt}] coordinates { 
(1, 38.55)
(2, 36.36)
(3, 36.97)
(4, 38.38)
(5, 43.03)
(6, 37.37)
(7, 37.27)
(8, 36.77)
(9, 41.11)
};
                \addlegendentry{\scalebox{.8}{Qwen2-Audio\ }}    
                
                \addplot[blue!60,mark=pentagon*,mark size=1.5pt,thick,mark options={fill=white,draw=blue,line width=0.5pt}] coordinates {
(1, 33.60)
(2, 35.45)
(3, 35.45)
(4, 34.75)
(5, 31.31)
(6, 33.03)
(7, 34.14)
(8, 30.81)
(9, 28.67)
};
                \addlegendentry{\scalebox{.8}{Mullama\ }}	
                \addplot[teal!70,mark=diamond*,mark size=1.5pt,thick,mark options={fill=white,draw=teal,line width=0.5pt}] coordinates {
(1,29.29)
(2,23.03)
(3,31.82)
(4,12.83)
(5,4.44)
(6,28.89)
(7,19.90)
(8,6.57)
(9,26.70)
                };
                \addlegendentry{\scalebox{.8}{Qwen-Audio\ }}    
                \addplot[red!60,mark=square*,mark size=1.5pt,thick,mark options={fill=white,draw=red,line width=0.5pt}] coordinates {
(1,0.2)
(2,0.4)
(3,4.85)
(4,0.0)
(5,0.2)
(6,0.1)
(7,0.0)
(8,0.3)
(9,0.4)
};
                \addlegendentry{\scalebox{.8}{GAMA\ }}    
                \addplot[gray!80,thick] coordinates {(0,33.3) (10,33.3)};
		\end{axis}
	}
	\scriptsize{
		\begin{axis}[
			at={(13em,0)},
			ymajorgrids,
			xmajorgrids,
			grid style=dashed,		
			legend style={at={(0.0,1.10)}, anchor=south west},
			legend columns=-1,
			xlabel={\scriptsize{\#Instruction}},
			xlabel style={yshift=1.2em},
			ylabel style={yshift=0em},xlabel style={yshift=0.0em},
			yticklabel style={/pgf/number format/precision=1,/pgf/number format/fixed zerofill},
			height=.26\textwidth,
			width=.29\textwidth,
			xmin=0,xmax=10,xtick={2,4,6,8},
			legend style={yshift=8pt,xshift=0em, legend plot pos=right,cells={anchor=west}}
			]
			\addplot[orange!80,mark=triangle*,,mark size=1.5pt,thick,mark options={fill=white,draw=orange,line width=0.5pt}] coordinates {

            };
					\addplot[orange!80,mark=triangle*,mark size=1.5pt,thick,mark options={fill=white,draw=orange,line width=0.5pt}] coordinates { 
(1, 27.67)
(2, 35.68)
(3, 13.73)
(4, 9.66)
(5, 9.95)
(6, 28.29)
(7, 21.87)
(8, 5.23)
(9, 18.92)
};
                
                \addplot[blue!60,mark=pentagon*,mark size=1.5pt,thick,mark options={fill=white,draw=blue,line width=0.5pt}] coordinates {
(1, 5.07)
(2, 1.91)
(3, 5.91)
(4, 0)
(5, 0)
(6, 1.87)
(7, 2.02)
(8, 1.8)
(9, 6.31)
};
                \addplot[teal!70,mark=diamond*,mark size=1.5pt,thick,mark options={fill=white,draw=teal,line width=0.5pt}] coordinates {
(1,18.84)
(2,13.05)
(3,8.97)
(4,4.29)
(5,5.43)
(6,13.95)
(7,15.32)
(8,5.37)
(9,9.62)
                };   
                \addplot[red!60,mark=square*,mark size=1.5pt,thick,mark options={fill=white,draw=red,line width=0.5pt}] coordinates {
(1,12.05)
(2,0.0)
(3,0.36)
(4,0.94)
(5,1.87)
(6,0.54)
(7,0.25)
(8,0)
(9,4.42)
};
                \addplot[gray!80,thick] coordinates {(0,11.1) (10,11.1)};	
		\end{axis}
	}
 
\end{tikzpicture}
	\caption{Performance of speech LLMs with different instructions on speaker age task (left) and scene classification task (right). Gray line shows random selection accuracy. Details about the  instructions and results are shown in App.~\ref{appendix:prompt}.}
	\label{instrction}
\end{wrapfigure}
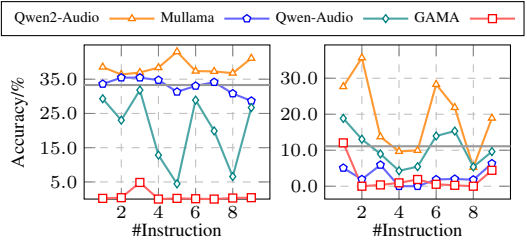
We classify the cause by observing changes in performance after perturbing the prompt. If the model is insensitive to different perturbed prompts, it indicates that the model cannot understand the prompt. On the other hand, if the models show significantly better performance with a properly structured prompt, it suggests that the model could understand the task, while requires the specific instruction. We choose the two Level \ref{eq:level3} tasks (Age prediction and Ambient Noise Detection) where the instruction following ability is crucial, and the results shown in Fig.~\ref{instrction}.

For the result of Fig.~\ref{instrction}, we can find the Mullama is not sensitive about the instruction. This prove the model can not figure out this task. Further, the performance of most speech LLMs highly related with the specific prompt, this shows models are sensitive with the format of instruction. Comparing with the text LLMs which are robust with diverse instruction, the speech LLMs need much effect to guarantee instruction following.
\begin{takeaway}
\textit{\textbf{Current speech LLMs follow instructions poorly.}}
\end{takeaway}

\subsection{Weak LLM Backbones}
\label{foundation LLMs}
\begin{table}[th]
    \centering
    \footnotesize
    \caption{Three tasks to evaluate the ability to process phonemes}
    \begin{tabular}{ll} 
        \toprule
        \textbf{Task} & \textbf{Prompt} \\
        \midrule
        Sequence-level & Given a phone sequence, ``M AA0 R K IH0 Z ...'', what sentence does it represent?\\
        \hline
        \multirow{2}{*}{Token-level} &Given a tokenized phone sequence, ``[M AA0 R K] [IH0 Z] ...'', what sentence \\ 
        & does it represent?  \\
        \hline
        \multirow{3}{*}{\makecell[l]{Token-level\\with one shot}} & Given a tokenized phone sequence, ``[M AA0 R K] [IH0 Z] ...'', what sentence  \\  & does it represent? For example, if the phone sequence is ``[F AO0 R] [F AY0 V], \\ &  [S IH0 K S] [S EH1 V N] [EY0 T]'' the sentence can be: ``four five six seven eight nine''. \\ 
        \bottomrule
    \end{tabular}
    \label{phone_prompt}
\end{table}

Most current speech LLMs follow the paradigm of stacking the acoustic model and text LLMs. This paradigm requires the text LLMs to process audio-like tokens, raising an intuitive question: whether text LLMs have the potential to handle cross-modal tasks. We designed a direct task of converting a phoneme sequence into a complete sentence. The phoneme represents pronunciation in text format, thus understanding phonemes can demonstrate the model's potential to process audio. We designed three different tasks, as shown in Tab.~\ref{phone_prompt}. The most challenging task requires the model to find the proper sentence according to the entire phoneme sequence, which takes some time even for humans.

\begin{wraptable}[15]{r}{0.5\textwidth}
\scriptsize
    \vspace{-10pt}
\footnotesize
    \centering
    \caption{Results of LLMs processing phonemes}

    \begin{tabular}{lccc} 

        \toprule
        \multirow{2}{*}{\textbf{Model}} & \multirow{2}{*}{\textbf{Seq.} $\downarrow$} & \textbf{Token} $\downarrow$& \textbf{Token} $\downarrow$\\
        &&zero-shot&one-shot\\
        \midrule
        GPT-4o & 17.5&	\ \ 8.3	&\ \ 8.3 \\
        \hline
        Mixtra-7B & 99.5&98.9&97.7\\
        Qwen2-7B &99.3&98.3&95.8 \\
        Llama3-7B&97.5&89.6&87.9\\
        Llama3.1-8B & 94.0&83.7&78.0\\
        \hline
        Mixtra 8x7B & 98.2&95.1&92.6\\
        Qwen2-72B &93.4 &75.4&73.5\\
        Llama3.1-70B & 80.5&51.1&46.9\\
        
        \bottomrule
    \end{tabular}
    \label{text-only_LLMS}
    The results are assessed using the WER. In instances where LLMs generate hallucinations or decline to provide a response, the WER is recorded as 100\%.
\end{wraptable}

We evaluate the most commonly used LLMs for building speech LLMs, and the results are shown in Tab.~\ref{text-only_LLMS}. We found that the closed-source GPT-4o demonstrates a surprising ability to process phonemes, proving that it can easily be converted into a powerful speech LLM. On the other hand, all the open-source models fail to show potential in handling audio. Even when the size of the model parameters is increased, the ability remains very limited.

One explanation is that open-source models overlook potential audio-related tasks, which is quite unlike GPT-4o. This leads to a significant gap between the two types of models. A piece of evidence supporting this is that Llama 3.1, which emphasizes multi-modal capabilities \cite{dubey2024llama}, shows a noticeable improvement in WER in token-level tasks and delivers robust performance with 70B parameters. Overall, open-source foundation models still have substantial room for improvement in their ability to handle audio-related tasks.

 \begin{takeaway}
    
 \textit{\textbf{The used LLM backbone is relatively weak for current speech LLMs.}}
\end{takeaway}

\section{Related Work}
\label{gen_inst}

Speech language models have seen a surge in development following the advent of LLMs. Currently, most work integrates pre-trained acoustic models with LLMs using an alignment module. There are two main strategies to bridge the gap between the two models: 1) adapters and 2) attention mechanisms. 

\textbf{Adapter}~ The former method adds modules (usually convolutional networks and MLPs) between the acoustic model and LLMs. Convolutional networks can compress sequence length ~\citep{wang2023blsp}, while MLPs are used to align acoustic tokens with text embeddings ~\citep{su2023pandagpt}. 

\textbf{Attention Mechanisms}~ Regarding the attention method, \cite{kong2024audio} implemented cross-attention to filter information from the output of the speech encoder. \cite{li2023blip} proposed the Q-former as an intermediate extractor based on cross-attention. Similarly, \cite{pan2023cosmic} applied the Q-former to extract useful acoustic information for LLMs. Some works directly treat the acoustic codec as tokens and do not rely on alignment strategies ~\citep{zhang2023speechgpt, rubenstein2023audiopalm}. 

\textbf{Categorization of speech LLMs}~
We have introduced that acoustic models can generally be divided into four types. Some works aim to build \textbf{universal multi-modal LLMs} ~\citep{su2023pandagpt, zhan2024anygpt, wu2023next, lyu2023macaw, zhang2023video, shukor2023unified}. Several studies focus on enhancing \textbf{music understanding}, an important area that has not yet received enough attention ~\citep{deshmukh2023pengi, zhan2024anygpt, liu2024music}. Most speech LLMs aim to improve \textbf{speech-to-text tasks} and \textbf{multi-turn dialogue capabilities} ~\citep{fathullah2024audiochatllama, shu2023llasm, wang2023slm, pan2023cosmic, rubenstein2023audiopalm, zhang2023speechgpt, bai2024seed, wu2023decoder, maiti2024voxtlm, wang2023blsp, qwen2audio, dubey2024llama}. Some works utilize audio codec models to enhance audio processing performance ~\citep{chen2023lauragpt, kong2024audio, nguyen2024spirit, das2024speechverse, gong2023joint}. Inspired by these efforts, several studies ~\citep{tang2023salmonn, ghosh2024gama, hu2024wavllm} combine acoustic and semantic codecs to integrate audio and speech processing capabilities into a single model.

\section{Conclusion}

In this paper, we explored the evolving landscape of large language models (LLMs) in the realm of speech processing. We introduced a five-level roadmap to guide the development of human-level speech understanding, from basic ASR capabilities to advanced generalist models that integrate non-semantic information with general abstract acoustic knowledge for complex tasks. To assess the current state of speech LLMs, we designed a comprehensive benchmark that standardizes critical aspects across various tasks, ensuring consistency and reliability in performance evaluation. Our research reveals the current stage and deficiencies in understanding speech by both humans and speech LLMs. We evaluate the advanced speech mode of GPT-4o and find that following speech instructions is very challenging. Further analysis has uncovered structural flaws in existing speech LLMs. Reveals that current speech LLMs face issues in both Acoustic Information Transfer and Foundation LLMs' Potentiality. The contributions of this paper provide a structured approach to advancing speech LLMs, offering valuable insights for future innovations in this field.

\bibliography{iclr2025_conference}

\begin{thebibliography}{60}
\providecommand{\natexlab}[1]{#1}
\providecommand{\url}[1]{\texttt{#1}}
\expandafter\ifx\csname urlstyle\endcsname\relax
  \providecommand{\doi}[1]{doi: #1}\else
  \providecommand{\doi}{doi: \begingroup \urlstyle{rm}\Url}\fi

\bibitem[Ao et~al.(2024)Ao, Wang, Tian, Chen, Zhang, Lu, Wang, Li, and
  Wu]{sdeval}
Junyi Ao, Yuancheng Wang, Xiaohai Tian, Dekun Chen, Jun Zhang, Lu~Lu, Yuxuan
  Wang, Haizhou Li, and Zhizheng Wu.
\newblock Sd-eval: A benchmark dataset for spoken dialogue understanding beyond
  words.
\newblock \emph{arXiv preprint arXiv:2406.13340}, 2024.

\bibitem[Bai et~al.(2024)Bai, Chen, Chen, Chen, Chen, Ding, Dong, Dong, Du,
  Gao, et~al.]{bai2024seed}
Ye~Bai, Jingping Chen, Jitong Chen, Wei Chen, Zhuo Chen, Chen Ding, Linhao
  Dong, Qianqian Dong, Yujiao Du, Kepan Gao, et~al.
\newblock Seed-asr: Understanding diverse speech and contexts with llm-based
  speech recognition.
\newblock \emph{arXiv preprint arXiv:2407.04675}, 2024.

\bibitem[Bapna et~al.(2021)Bapna, Chung, Wu, Gulati, Jia, Clark, Johnson,
  Riesa, Conneau, and Zhang]{bapna2021slam}
Ankur Bapna, Yu-an Chung, Nan Wu, Anmol Gulati, Ye~Jia, Jonathan~H Clark,
  Melvin Johnson, Jason Riesa, Alexis Conneau, and Yu~Zhang.
\newblock Slam: A unified encoder for speech and language modeling via
  speech-text joint pre-training.
\newblock \emph{arXiv preprint arXiv:2110.10329}, 2021.

\bibitem[Chaudhari et~al.(2020)Chaudhari, Jiang, Fakhry, Han, Xiao, Shen, and
  Khanzada]{virufy}
Gunvant Chaudhari, Xinyi Jiang, Ahmed Fakhry, Asriel Han, Jaclyn Xiao, Sabrina
  Shen, and Amil Khanzada.
\newblock Virufy: Global applicability of crowdsourced and clinical datasets
  for ai detection of covid-19 from cough.
\newblock \emph{arXiv preprint arXiv:2011.13320}, 2020.

\bibitem[Chen et~al.(2023)Chen, Chu, Gao, Li, Hu, Zhou, Xu, Ma, Wang, Zheng,
  et~al.]{chen2023lauragpt}
Qian Chen, Yunfei Chu, Zhifu Gao, Zerui Li, Kai Hu, Xiaohuan Zhou, Jin Xu,
  Ziyang Ma, Wen Wang, Siqi Zheng, et~al.
\newblock Lauragpt: Listen, attend, understand, and regenerate audio with gpt.
\newblock \emph{arXiv preprint arXiv:2310.04673}, 2023.

\bibitem[Chien et~al.(2021)Chien, Lin, Huang, Hsu, and Lee]{mult-speaker}
Chung-Ming Chien, Jheng-Hao Lin, Chien-yu Huang, Po-chun Hsu, and Hung-yi Lee.
\newblock Investigating on incorporating pretrained and learnable speaker
  representations for multi-speaker multi-style text-to-speech.
\newblock In \emph{ICASSP 2021 - 2021 IEEE International Conference on
  Acoustics, Speech and Signal Processing (ICASSP)}, pp.\  8588--8592, 2021.
\newblock \doi{10.1109/ICASSP39728.2021.9413880}.

\bibitem[Chu et~al.(2023)Chu, Xu, Zhou, Yang, Zhang, Yan, Zhou, and
  Zhou]{qwen_audio}
Yunfei Chu, Jin Xu, Xiaohuan Zhou, Qian Yang, Shiliang Zhang, Zhijie Yan, Chang
  Zhou, and Jingren Zhou.
\newblock Qwen-audio: Advancing universal audio understanding via unified
  large-scale audio-language models.
\newblock \emph{arXiv preprint arXiv:2311.07919}, 2023.

\bibitem[Chu et~al.(2024)Chu, Xu, Yang, Wei, Wei, Guo, Leng, Lv, He, Lin,
  et~al.]{qwen2audio}
Yunfei Chu, Jin Xu, Qian Yang, Haojie Wei, Xipin Wei, Zhifang Guo, Yichong
  Leng, Yuanjun Lv, Jinzheng He, Junyang Lin, et~al.
\newblock Qwen2-audio technical report.
\newblock \emph{arXiv preprint arXiv:2407.10759}, 2024.

\bibitem[C\'ifka et~al.(2023)C\'ifka, Dimitriou, Wang, Schreiber, Miner, and
  St\"oter]{jam-alt}
Ond\v{r}ej C\'ifka, Constantinos Dimitriou, {Cheng-i} Wang, Hendrik Schreiber,
  Luke Miner, and Fabian-Robert St\"oter.
\newblock {Jam-ALT}: A formatting-aware lyrics transcription benchmark.
\newblock \emph{arXiv preprint arXiv:2311.13987}, 2023.

\bibitem[Das et~al.(2024)Das, Dingliwal, Ronanki, Paturi, Huang, Mathur, Yuan,
  Bekal, Niu, Jayanthi, et~al.]{das2024speechverse}
Nilaksh Das, Saket Dingliwal, Srikanth Ronanki, Rohit Paturi, David Huang,
  Prashant Mathur, Jie Yuan, Dhanush Bekal, Xing Niu, Sai~Muralidhar Jayanthi,
  et~al.
\newblock Speechverse: A large-scale generalizable audio language model.
\newblock \emph{arXiv preprint arXiv:2405.08295}, 2024.

\bibitem[Deshmukh et~al.(2023)Deshmukh, Elizalde, Singh, and
  Wang]{deshmukh2023pengi}
Soham Deshmukh, Benjamin Elizalde, Rita Singh, and Huaming Wang.
\newblock Pengi: An audio language model for audio tasks.
\newblock \emph{Advances in Neural Information Processing Systems},
  36:\penalty0 18090--18108, 2023.

\bibitem[Dubey et~al.(2024)Dubey, Jauhri, Pandey, Kadian, Al-Dahle, Letman,
  Mathur, Schelten, Yang, Fan, et~al.]{dubey2024llama}
Abhimanyu Dubey, Abhinav Jauhri, Abhinav Pandey, Abhishek Kadian, Ahmad
  Al-Dahle, Aiesha Letman, Akhil Mathur, Alan Schelten, Amy Yang, Angela Fan,
  et~al.
\newblock The llama 3 herd of models.
\newblock \emph{arXiv preprint arXiv:2407.21783}, 2024.

\bibitem[Durand et~al.(2023)Durand, Stoller, and Ewert]{Jam-Lyrics}
Simon Durand, Daniel Stoller, and Sebastian Ewert.
\newblock Contrastive learning-based audio to lyrics alignment for multiple
  languages.
\newblock In \emph{{IEEE} International Conference on Acoustics, Speech and
  Signal Processing {ICASSP} 2023, Rhodes Island, Greece, June 4-10, 2023},
  pp.\  1--5. {IEEE}, 2023.
\newblock \doi{10.1109/ICASSP49357.2023.10096725}.

\bibitem[Fathullah et~al.(2024)Fathullah, Wu, Lakomkin, Li, Jia, Shangguan,
  Mahadeokar, Kalinli, Fuegen, and Seltzer]{fathullah2024audiochatllama}
Yassir Fathullah, Chunyang Wu, Egor Lakomkin, Ke~Li, Junteng Jia, Yuan
  Shangguan, Jay Mahadeokar, Ozlem Kalinli, Christian Fuegen, and Mike Seltzer.
\newblock Audiochatllama: Towards general-purpose speech abilities for llms.
\newblock In \emph{Proceedings of the 2024 Conference of the North American
  Chapter of the Association for Computational Linguistics: Human Language
  Technologies (Volume 1: Long Papers)}, pp.\  5522--5532, 2024.

\bibitem[Gao et~al.(2024)Gao, Chen, Du, Xu, Guo, Bu, Yang, Li, and
  Lee]{aishellMDSC}
Ming Gao, Hang Chen, Jun Du, Xin Xu, Hongxiao Guo, Hui Bu, Jianxing Yang, Ming
  Li, and Chin-Hui Lee.
\newblock Enhancing voice wake-up for dysarthria: Mandarin dysarthria speech
  corpus release and customized system design.
\newblock \emph{arXiv preprint arXiv:2406.10304}, 2024.

\bibitem[Ghosh et~al.(2024{\natexlab{a}})Ghosh, Kumar, Seth, Evuru, Tyagi,
  Sakshi, Nieto, Duraiswami, and Manocha]{ghosh2024gama}
Sreyan Ghosh, Sonal Kumar, Ashish Seth, Chandra Kiran~Reddy Evuru, Utkarsh
  Tyagi, S~Sakshi, Oriol Nieto, Ramani Duraiswami, and Dinesh Manocha.
\newblock Gama: A large audio-language model with advanced audio understanding
  and complex reasoning abilities.
\newblock \emph{arXiv preprint arXiv:2406.11768}, 2024{\natexlab{a}}.

\bibitem[Ghosh et~al.(2024{\natexlab{b}})Ghosh, Kumar, Seth, Evuru, Tyagi,
  Singh, Nieto, Duraiswami, and Manocha]{GAMA}
Sreyan Ghosh, Sonal Kumar, Ashish Seth, Chandra Kiran~Reddy Evuru, Utkarsh
  Tyagi, Sakshi Singh, Oriol Nieto, Ramani Duraiswami, and Dinesh Manocha.
\newblock {GAMA:} {A} large audio-language model with advanced audio
  understanding and complex reasoning abilities.
\newblock \emph{arXiv preprint arXiv:2406.11768}, 2024{\natexlab{b}}.

\bibitem[Gong et~al.(2023)Gong, Liu, Luo, Karlinsky, and Glass]{gong2023joint}
Yuan Gong, Alexander~H Liu, Hongyin Luo, Leonid Karlinsky, and James Glass.
\newblock Joint audio and speech understanding.
\newblock In \emph{2023 IEEE Automatic Speech Recognition and Understanding
  Workshop (ASRU)}, pp.\  1--8. IEEE, 2023.

\bibitem[Hu et~al.(2024)Hu, Zhou, Liu, Chen, Hao, Pan, Liu, Li, Sivasankaran,
  Liu, et~al.]{hu2024wavllm}
Shujie Hu, Long Zhou, Shujie Liu, Sanyuan Chen, Hongkun Hao, Jing Pan, Xunying
  Liu, Jinyu Li, Sunit Sivasankaran, Linquan Liu, et~al.
\newblock Wavllm: Towards robust and adaptive speech large language model.
\newblock \emph{arXiv preprint arXiv:2404.00656}, 2024.

\bibitem[Huang et~al.(2024)Huang, Lu, Wang, Hsiao, Kuan, Wu, Arora, Chang, Shi,
  Peng, et~al.]{huang2024dynamic}
Chien-yu Huang, Ke-Han Lu, Shih-Heng Wang, Chi-Yuan Hsiao, Chun-Yi Kuan, Haibin
  Wu, Siddhant Arora, Kai-Wei Chang, Jiatong Shi, Yifan Peng, et~al.
\newblock Dynamic-superb: Towards a dynamic, collaborative, and comprehensive
  instruction-tuning benchmark for speech.
\newblock In \emph{ICASSP 2024-2024 IEEE International Conference on Acoustics,
  Speech and Signal Processing (ICASSP)}, pp.\  12136--12140. IEEE, 2024.

\bibitem[Iranzo{-}S{\'{a}}nchez et~al.(2020)Iranzo{-}S{\'{a}}nchez,
  Silvestre{-}Cerd{\`{a}}, Jorge, Rosell{\'{o}}, Gim{\'{e}}nez, Sanch{\'{\i}}s,
  Civera, and Juan]{Europarl-ST}
Javier Iranzo{-}S{\'{a}}nchez, Joan~Albert Silvestre{-}Cerd{\`{a}}, Javier
  Jorge, Nahuel Rosell{\'{o}}, Adri{\`{a}} Gim{\'{e}}nez, Albert
  Sanch{\'{\i}}s, Jorge Civera, and Alfons Juan.
\newblock Europarl-st: {A} multilingual corpus for speech translation of
  parliamentary debates.
\newblock In \emph{2020 {IEEE} International Conference on Acoustics, Speech
  and Signal Processing, {ICASSP} 2020, Barcelona, Spain, May 4-8, 2020}, pp.\
  8229--8233. {IEEE}, 2020.
\newblock \doi{10.1109/ICASSP40776.2020.9054626}.

\bibitem[Ito \& Johnson(2017)Ito and Johnson]{ljspeech}
Keith Ito and Linda Johnson.
\newblock The lj speech dataset.
\newblock \url{https://keithito.com/LJ-Speech-Dataset/}, 2017.

\bibitem[Kong et~al.(2024)Kong, Goel, Badlani, Ping, Valle, and
  Catanzaro]{kong2024audio}
Zhifeng Kong, Arushi Goel, Rohan Badlani, Wei Ping, Rafael Valle, and Bryan
  Catanzaro.
\newblock Audio flamingo: A novel audio language model with few-shot learning
  and dialogue abilities.
\newblock \emph{arXiv preprint arXiv:2402.01831}, 2024.

\bibitem[Krathwohl(2002)]{krathwohl2002revision}
David~R Krathwohl.
\newblock A revision of bloom's taxonomy: An overview.
\newblock \emph{Theory into practice}, 41\penalty0 (4):\penalty0 212--218,
  2002.

\bibitem[Li et~al.(2023)Li, Li, Savarese, and Hoi]{li2023blip}
Junnan Li, Dongxu Li, Silvio Savarese, and Steven Hoi.
\newblock Blip-2: Bootstrapping language-image pre-training with frozen image
  encoders and large language models.
\newblock In \emph{International conference on machine learning}, pp.\
  19730--19742. PMLR, 2023.

\bibitem[Liu et~al.(2023)Liu, Li, Wu, and Lee]{liu2023llava}
Haotian Liu, Chunyuan Li, Qingyang Wu, and Yong~Jae Lee.
\newblock Visual instruction tuning, 2023.

\bibitem[Liu et~al.(2024{\natexlab{a}})Liu, Hussain, Sun, and
  Shan]{liu2024music}
Shansong Liu, Atin~Sakkeer Hussain, Chenshuo Sun, and Ying Shan.
\newblock Music understanding llama: Advancing text-to-music generation with
  question answering and captioning.
\newblock In \emph{ICASSP 2024-2024 IEEE International Conference on Acoustics,
  Speech and Signal Processing (ICASSP)}, pp.\  286--290. IEEE,
  2024{\natexlab{a}}.

\bibitem[Liu et~al.(2024{\natexlab{b}})Liu, Hussain, Sun, and Shan]{mullama}
Shansong Liu, Atin~Sakkeer Hussain, Chenshuo Sun, and Ying Shan.
\newblock Music understanding llama: Advancing text-to-music generation with
  question answering and captioning.
\newblock In \emph{{IEEE} International Conference on Acoustics, Speech and
  Signal Processing, {ICASSP} 2024, Seoul, Republic of Korea, April 14-19,
  2024}, pp.\  286--290. {IEEE}, 2024{\natexlab{b}}.
\newblock \doi{10.1109/ICASSP48485.2024.10447027}.

\bibitem[Livingstone \& Russo(2018)Livingstone and Russo]{ravdess}
Steven~R Livingstone and Frank~A Russo.
\newblock The ryerson audio-visual database of emotional speech and song
  (ravdess): A dynamic, multimodal set of facial and vocal expressions in north
  american english.
\newblock \emph{PloS one}, 13\penalty0 (5):\penalty0 e0196391, 2018.

\bibitem[Lyu et~al.(2023)Lyu, Wu, Wang, Huang, Liu, Du, Shi, and
  Tu]{lyu2023macaw}
Chenyang Lyu, Minghao Wu, Longyue Wang, Xinting Huang, Bingshuai Liu, Zefeng
  Du, Shuming Shi, and Zhaopeng Tu.
\newblock Macaw-llm: Multi-modal language modeling with image, audio, video,
  and text integration.
\newblock \emph{arXiv preprint arXiv:2306.09093}, 2023.

\bibitem[Maiti et~al.(2024)Maiti, Peng, Choi, Jung, Chang, and
  Watanabe]{maiti2024voxtlm}
Soumi Maiti, Yifan Peng, Shukjae Choi, Jee-weon Jung, Xuankai Chang, and Shinji
  Watanabe.
\newblock Voxtlm: Unified decoder-only models for consolidating speech
  recognition, synthesis and speech, text continuation tasks.
\newblock In \emph{ICASSP 2024-2024 IEEE International Conference on Acoustics,
  Speech and Signal Processing (ICASSP)}, pp.\  13326--13330. IEEE, 2024.

\bibitem[Nguyen et~al.(2024)Nguyen, Muller, Yu, Costa-Jussa, Elbayad, Popuri,
  Duquenne, Algayres, Mavlyutov, Gat, et~al.]{nguyen2024spirit}
Tu~Anh Nguyen, Benjamin Muller, Bokai Yu, Marta~R Costa-Jussa, Maha Elbayad,
  Sravya Popuri, Paul-Ambroise Duquenne, Robin Algayres, Ruslan Mavlyutov, Itai
  Gat, et~al.
\newblock Spirit-lm: Interleaved spoken and written language model.
\newblock \emph{arXiv preprint arXiv:2402.05755}, 2024.

\bibitem[OpenAI(2023)]{OpenAI2023GPT4}
OpenAI.
\newblock Gpt-4: Largest language model ever with 100 trillion parameters,
  2023.
\newblock URL \url{https://openai.com/blog/gpt-4/}.
\newblock Accessed: 2023-04-01.

\bibitem[Orlandic et~al.(2021)Orlandic, Teijeiro, and Atienza]{coughvid}
Lara Orlandic, Tomas Teijeiro, and David Atienza.
\newblock The coughvid crowdsourcing dataset, a corpus for the study of
  large-scale cough analysis algorithms.
\newblock \emph{Scientific Data}, 8\penalty0 (1):\penalty0 156, 2021.

\bibitem[Pan et~al.(2023)Pan, Wu, Gaur, Sivasankaran, Chen, Liu, and
  Li]{pan2023cosmic}
Jing Pan, Jian Wu, Yashesh Gaur, Sunit Sivasankaran, Zhuo Chen, Shujie Liu, and
  Jinyu Li.
\newblock Cosmic: Data efficient instruction-tuning for speech in-context
  learning.
\newblock \emph{arXiv preprint arXiv:2311.02248}, 2023.

\bibitem[Panayotov et~al.(2015)Panayotov, Chen, Povey, and
  Khudanpur]{LibriSpeech}
Vassil Panayotov, Guoguo Chen, Daniel Povey, and Sanjeev Khudanpur.
\newblock Librispeech: An {ASR} corpus based on public domain audio books.
\newblock In \emph{2015 {IEEE} International Conference on Acoustics, Speech
  and Signal Processing, {ICASSP} 2015, South Brisbane, Queensland, Australia,
  April 19-24, 2015}, pp.\  5206--5210. {IEEE}, 2015.
\newblock \doi{10.1109/ICASSP.2015.7178964}.

\bibitem[Pratap et~al.(2024)Pratap, Tjandra, Shi, Tomasello, Babu, Kundu,
  Elkahky, Ni, Vyas, Fazel-Zarandi, et~al.]{pratap2024scaling}
Vineel Pratap, Andros Tjandra, Bowen Shi, Paden Tomasello, Arun Babu, Sayani
  Kundu, Ali Elkahky, Zhaoheng Ni, Apoorv Vyas, Maryam Fazel-Zarandi, et~al.
\newblock Scaling speech technology to 1,000+ languages.
\newblock \emph{Journal of Machine Learning Research}, 25\penalty0
  (97):\penalty0 1--52, 2024.

\bibitem[Radford et~al.(2023)Radford, Kim, Xu, Brockman, McLeavey, and
  Sutskever]{whisper}
Alec Radford, Jong~Wook Kim, Tao Xu, Greg Brockman, Christine McLeavey, and
  Ilya Sutskever.
\newblock Robust speech recognition via large-scale weak supervision.
\newblock In Andreas Krause, Emma Brunskill, Kyunghyun Cho, Barbara Engelhardt,
  Sivan Sabato, and Jonathan Scarlett (eds.), \emph{International Conference on
  Machine Learning, {ICML} 2023, 23-29 July 2023, Honolulu, Hawaii, {USA}},
  volume 202 of \emph{Proceedings of Machine Learning Research}, pp.\
  28492--28518. {PMLR}, 2023.
\newblock URL \url{https://proceedings.mlr.press/v202/radford23a.html}.

\bibitem[Reddy et~al.(2019)Reddy, Beyrami, Pool, Cutler, Srinivasan, and
  Gehrke]{MSSNSD}
Chandan~KA Reddy, Ebrahim Beyrami, Jamie Pool, Ross Cutler, Sriram Srinivasan,
  and Johannes Gehrke.
\newblock A scalable noisy speech dataset and online subjective test framework.
\newblock \emph{Proc. Interspeech 2019}, pp.\  1816--1820, 2019.

\bibitem[Rubenstein et~al.(2023)Rubenstein, Asawaroengchai, Nguyen, Bapna,
  Borsos, Quitry, Chen, Badawy, Han, Kharitonov,
  et~al.]{rubenstein2023audiopalm}
Paul~K Rubenstein, Chulayuth Asawaroengchai, Duc~Dung Nguyen, Ankur Bapna,
  Zal{\'a}n Borsos, F{\'e}lix de~Chaumont Quitry, Peter Chen, Dalia~El Badawy,
  Wei Han, Eugene Kharitonov, et~al.
\newblock Audiopalm: A large language model that can speak and listen.
\newblock \emph{arXiv preprint arXiv:2306.12925}, 2023.

\bibitem[{Seamless Communication} et~al.(2023){Seamless Communication},
  Barrault, Chung, Meglioli, Dale, Dong, Duppenthaler, Duquenne, Ellis,
  Elsahar, Haaheim, et~al.]{barrault2023seamless}
{Seamless Communication}, Lo{\"\i}c Barrault, Yu-An Chung, Mariano~Coria
  Meglioli, David Dale, Ning Dong, Mark Duppenthaler, Paul-Ambroise Duquenne,
  Brian Ellis, Hady Elsahar, Justin Haaheim, et~al.
\newblock Seamless: Multilingual expressive and streaming speech translation.
\newblock \emph{arXiv preprint arXiv:2312.05187}, 2023.

\bibitem[Shu et~al.(2023)Shu, Dong, Chen, Huang, Zhang, Shi, Xiang, and
  Shi]{shu2023llasm}
Yu~Shu, Siwei Dong, Guangyao Chen, Wenhao Huang, Ruihua Zhang, Daochen Shi,
  Qiqi Xiang, and Yemin Shi.
\newblock Llasm: Large language and speech model.
\newblock \emph{arXiv preprint arXiv:2308.15930}, 2023.

\bibitem[Shukor et~al.(2023)Shukor, Dancette, Rame, and
  Cord]{shukor2023unified}
Mustafa Shukor, Corentin Dancette, Alexandre Rame, and Matthieu Cord.
\newblock Unified model for image, video, audio and language tasks.
\newblock \emph{arXiv preprint arXiv:2307.16184}, 2023.

\bibitem[SpeechTeam(2024)]{funaudiollm}
Tongyi SpeechTeam.
\newblock Funaudiollm: Voice understanding and generation foundation models for
  natural interaction between humans and llms.
\newblock \emph{arXiv preprint arXiv:2407.04051}, 2024.

\bibitem[Su et~al.(2023)Su, Lan, Li, Xu, Wang, and Cai]{su2023pandagpt}
Yixuan Su, Tian Lan, Huayang Li, Jialu Xu, Yan Wang, and Deng Cai.
\newblock Pandagpt: One model to instruction-follow them all.
\newblock \emph{arXiv preprint arXiv:2305.16355}, 2023.

\bibitem[Tang et~al.(2023)Tang, Yu, Sun, Chen, Tan, Li, Lu, Ma, and
  Zhang]{tang2023salmonn}
Changli Tang, Wenyi Yu, Guangzhi Sun, Xianzhao Chen, Tian Tan, Wei Li, Lu~Lu,
  Zejun Ma, and Chao Zhang.
\newblock Salmonn: Towards generic hearing abilities for large language models.
\newblock \emph{arXiv preprint arXiv:2310.13289}, 2023.

\bibitem[Touvron et~al.(2023)Touvron, Martin, Stone, Albert, Almahairi, Babaei,
  Bashlykov, Batra, Bhargava, Bhosale, et~al.]{llama2}
Hugo Touvron, Louis Martin, Kevin Stone, Peter Albert, Amjad Almahairi, Yasmine
  Babaei, Nikolay Bashlykov, Soumya Batra, Prajjwal Bhargava, Shruti Bhosale,
  et~al.
\newblock Llama 2: Open foundation and fine-tuned chat models.
\newblock \emph{arXiv preprint arXiv:2307.09288}, 2023.

\bibitem[Valentini-Botinhao et~al.(2017)]{noisySpeech}
Cassia Valentini-Botinhao et~al.
\newblock Noisy speech database for training speech enhancement algorithms and
  tts models.
\newblock \emph{University of Edinburgh. School of Informatics. Centre for
  Speech Technology Research (CSTR)}, 2017.

\bibitem[Wang et~al.(2023{\natexlab{a}})Wang, Liao, Huang, Lu, Wu, Liu, Zong,
  and Zhang]{wang2023blsp}
Chen Wang, Minpeng Liao, Zhongqiang Huang, Jinliang Lu, Junhong Wu, Yuchen Liu,
  Chengqing Zong, and Jiajun Zhang.
\newblock Blsp: Bootstrapping language-speech pre-training via behavior
  alignment of continuation writing.
\newblock \emph{arXiv preprint arXiv:2309.00916}, 2023{\natexlab{a}}.

\bibitem[Wang et~al.(2023{\natexlab{b}})Wang, Han, Shafran, Wu, Chiu, Cao,
  Chen, Zhang, Soltau, Rubenstein, et~al.]{wang2023slm}
Mingqiu Wang, Wei Han, Izhak Shafran, Zelin Wu, Chung-Cheng Chiu, Yuan Cao,
  Nanxin Chen, Yu~Zhang, Hagen Soltau, Paul~K Rubenstein, et~al.
\newblock Slm: Bridge the thin gap between speech and text foundation models.
\newblock In \emph{2023 IEEE Automatic Speech Recognition and Understanding
  Workshop (ASRU)}, pp.\  1--8. IEEE, 2023{\natexlab{b}}.

\bibitem[Weinberger(2013)]{SpeechAccentArchive}
S.~Weinberger.
\newblock Speech accent archive, 2013.

\bibitem[Wu et~al.(2023{\natexlab{a}})Wu, Gaur, Chen, Zhou, Zhu, Wang, Li, Liu,
  Ren, Liu, et~al.]{wu2023decoder}
Jian Wu, Yashesh Gaur, Zhuo Chen, Long Zhou, Yimeng Zhu, Tianrui Wang, Jinyu
  Li, Shujie Liu, Bo~Ren, Linquan Liu, et~al.
\newblock On decoder-only architecture for speech-to-text and large language
  model integration.
\newblock In \emph{2023 IEEE Automatic Speech Recognition and Understanding
  Workshop (ASRU)}, pp.\  1--8. IEEE, 2023{\natexlab{a}}.

\bibitem[Wu et~al.(2023{\natexlab{b}})Wu, Fei, Qu, Ji, and Chua]{wu2023next}
Shengqiong Wu, Hao Fei, Leigang Qu, Wei Ji, and Tat-Seng Chua.
\newblock Next-gpt: Any-to-any multimodal llm.
\newblock \emph{arXiv preprint arXiv:2309.05519}, 2023{\natexlab{b}}.

\bibitem[Yamagishi et~al.(2019)Yamagishi, Veaux, and MacDonald]{vctk}
Junichi Yamagishi, Christophe Veaux, and Kirsten MacDonald.
\newblock {CSTR VCTK Corpus}: English multi-speaker corpus for {CSTR} voice
  cloning toolkit (version 0.92), 2019.

\bibitem[Yang et~al.(2024)Yang, Xu, Liu, Chu, Jiang, Zhou, Leng, Lv, Zhao,
  Zhou, et~al.]{airbench}
Qian Yang, Jin Xu, Wenrui Liu, Yunfei Chu, Ziyue Jiang, Xiaohuan Zhou, Yichong
  Leng, Yuanjun Lv, Zhou Zhao, Chang Zhou, et~al.
\newblock Air-bench: Benchmarking large audio-language models via generative
  comprehension.
\newblock \emph{arXiv preprint arXiv:2402.07729}, 2024.

\bibitem[Zhan et~al.(2024)Zhan, Dai, Ye, Zhou, Zhang, Liu, Zhang, Yuan, Zhang,
  Li, et~al.]{zhan2024anygpt}
Jun Zhan, Junqi Dai, Jiasheng Ye, Yunhua Zhou, Dong Zhang, Zhigeng Liu, Xin
  Zhang, Ruibin Yuan, Ge~Zhang, Linyang Li, et~al.
\newblock Anygpt: Unified multimodal llm with discrete sequence modeling.
\newblock \emph{arXiv preprint arXiv:2402.12226}, 2024.

\bibitem[Zhang et~al.(2023{\natexlab{a}})Zhang, Li, Zhang, Zhan, Wang, Zhou,
  and Qiu]{zhang2023speechgpt}
Dong Zhang, Shimin Li, Xin Zhang, Jun Zhan, Pengyu Wang, Yaqian Zhou, and
  Xipeng Qiu.
\newblock Speechgpt: Empowering large language models with intrinsic
  cross-modal conversational abilities.
\newblock \emph{arXiv preprint arXiv:2305.11000}, 2023{\natexlab{a}}.

\bibitem[Zhang et~al.(2023{\natexlab{b}})Zhang, Li, and Bing]{zhang2023video}
Hang Zhang, Xin Li, and Lidong Bing.
\newblock Video-llama: An instruction-tuned audio-visual language model for
  video understanding.
\newblock \emph{arXiv preprint arXiv:2306.02858}, 2023{\natexlab{b}}.

\bibitem[Zhang et~al.(2021)Zhang, Zhang, Wang, Yan, Song, Huang, Li, Povey, and
  Wang]{speechocean762}
Junbo Zhang, Zhiwen Zhang, Yongqing Wang, Zhiyong Yan, Qiong Song, Yukai Huang,
  Ke~Li, Daniel Povey, and Yujun Wang.
\newblock speechocean762: An open-source non-native english speech corpus for
  pronunciation assessment.
\newblock In Hynek Hermansky, Honza Cernock{\'{y}}, Luk{\'{a}}s Burget, Lori
  Lamel, Odette Scharenborg, and Petr Motl{\'{\i}}cek (eds.), \emph{22nd Annual
  Conference of the International Speech Communication Association, Interspeech
  2021, Brno, Czechia, August 30 - September 3, 2021}, pp.\  3710--3714.
  {ISCA}, 2021.
\newblock \doi{10.21437/INTERSPEECH.2021-1259}.
\newblock URL \url{https://doi.org/10.21437/Interspeech.2021-1259}.

\bibitem[Zhang et~al.(2023{\natexlab{c}})Zhang, Han, Qin, Wang, Bapna, Chen,
  Chen, Li, Axelrod, Wang, et~al.]{zhang2023google}
Yu~Zhang, Wei Han, James Qin, Yongqiang Wang, Ankur Bapna, Zhehuai Chen, Nanxin
  Chen, Bo~Li, Vera Axelrod, Gary Wang, et~al.
\newblock Google usm: Scaling automatic speech recognition beyond 100
  languages.
\newblock \emph{arXiv preprint arXiv:2303.01037}, 2023{\natexlab{c}}.

\end{thebibliography}
\bibliographystyle{iclr2025_conference}

\newpage
\appendix
\newpage
\section*{Limitation}
Artificial intelligence should not be confined to overly narrow domains, as such a focus can lead to frequent model switching when handling diverse tasks.This requires SAGI, a speech AGI, to be a powerful assistant capable of completing all kinds of tasks. However, during our primary testing, most speech LLMs remain at levels 1 and 2, indicating there is still a long way to go in terms of understanding speech.

To advance further, we conclude some important directions for improving speech LLMs toward higher level:
\begin{itemize}
    \item Requiring more diverse speech data to handle complex tasks.
    \item Enhancing the ability of text LLMs to process speech-related tasks.
    \item Ensuring that LLMs can receive complete acoustic information.
\end{itemize}
We advocate for the development of more powerful acoustic models, consideration of cross-domain compatibility when constructing datasets, and a deepening of expertise in specific research areas. This approach will enhance the generalization and adaptability of the models.

\section{Existing Benchmark}
\label{sec:existing}

Tab.~\ref{tab:existing_benchmarks} summarizes the coverage of existing benchmarks across different levels of speech model tasks, highlighting gaps in current evaluation methods. L1 tasks such as Speech ASR, Intent Classification, and Language Identification are well supported by both Dynamic-SUPERB and AIR-Bench, though SD-Eval \citep{sdeval} lacks coverage. For Level~\ref{eq:level2} foundational perception tasks, like Music Pitch and Velocity, only AIR-Bench \citep{airbench} provides support. Level~\ref{eq:level3} tasks related to non-semantic comprehension, such as Emotion, Environment, and Speaker Gender/Age, are covered to varying degrees across all benchmarks, with Dynamic-SUPERB \citep{huang2024dynamic} offering the most comprehensive support. However, more specialized tasks like Sarcasm, Stress, and Spoof Detection are only covered by Dynamic-SUPERB. Notably, Level~\ref{eq:level4} (Abstract Knowledge) and Level~\ref{eq:level5} (Speech AGI) remain entirely unsupported across all benchmarks. This underscores the urgent need to build a more comprehensive benchmark that addresses the gaps in Level~\ref{eq:level2}, Level~\ref{eq:level4}, and Level~\ref{eq:level5}, ensuring more robust evaluation across all levels of speech model tasks.

\begin{table}[h]
\label{tab:existing_benchmarks}
    \centering
    \small
    \caption{Existing benchmarks across Levels. L2, L4 and L5 have not received enough attention yet.}
    \begin{tabular}{clccc}
        \toprule
        \textbf{Level} & \textbf{Task} & \textbf{Dynamic-SUPERB} & \textbf{AIR-Bench} & \textbf{SD-Eval} \\ \midrule
        \multirow{3}{*}{L1} & Speech ASR & \cmark & \cmark & \xmark \\ 
                            & Intent Classification & \cmark & \cmark & \xmark \\ 
                            & Language Identification & \cmark & \cmark & \xmark \\ \hline
        \rowcolor{black!10} \multirow{1}{*}{L2} & Music Pitch and Velocity & \xmark & \cmark & \xmark \\ \hline
        \multirow{12}{*}{L3} & Emotion & \cmark & \cmark & \cmark \\ 
                            & Environment & \cmark & \cmark & \cmark \\ 
                            & Accent & \cmark & \xmark & \cmark \\ 
                            & Speaker Gender/Age & \xmark & \cmark & \cmark \\ 
                            & Noise Detection & \cmark & \xmark & \xmark \\ 
                            & Speaker Verification & \cmark & \cmark & \xmark \\ 
                            & Sarcasm Detection & \cmark & \xmark & \xmark \\ 
                            & Stress Detection & \cmark & \xmark & \xmark \\ 
                            & How Far Are You & \cmark & \xmark & \xmark \\ 
                            & Spoof Detection & \cmark & \xmark & \xmark \\ 
                            & Synthesized Voice Detection & \xmark & \cmark & \xmark \\ \hline
        \rowcolor{black!10} \multirow{1}{*}{L4} & No Related Work & \xmark & \xmark & \xmark \\ \hline
        \rowcolor{black!10} \multirow{1}{*}{L5} & No Related Work & \xmark & \xmark & \xmark \\ \bottomrule
    \end{tabular}
\end{table}

\section{Details of Benchmark Construction}
\label{appendix:benchmark-cons-details}
The overall construction principles are provided in Sec.~\ref{appendix:general-principle}. The data and tools used are detailed in Sec.~\ref{appendix:data_tool_use}. The composition structure of the data is outlined in Sec.~\ref{sec:benchmark_data_structure}. Detailed construction details for each task are available in Sec.~\ref{appendix:task_detail}.The credibility verification of synthesized speech is provided in Sec.~\ref{app:cred_verify}.

\subsection{General Principles of Data Construction}
\label{appendix:general-principle}
\subsubsection{Question Construction}
\begin{figure}[h]
    \centering
    \includegraphics[scale=0.28]{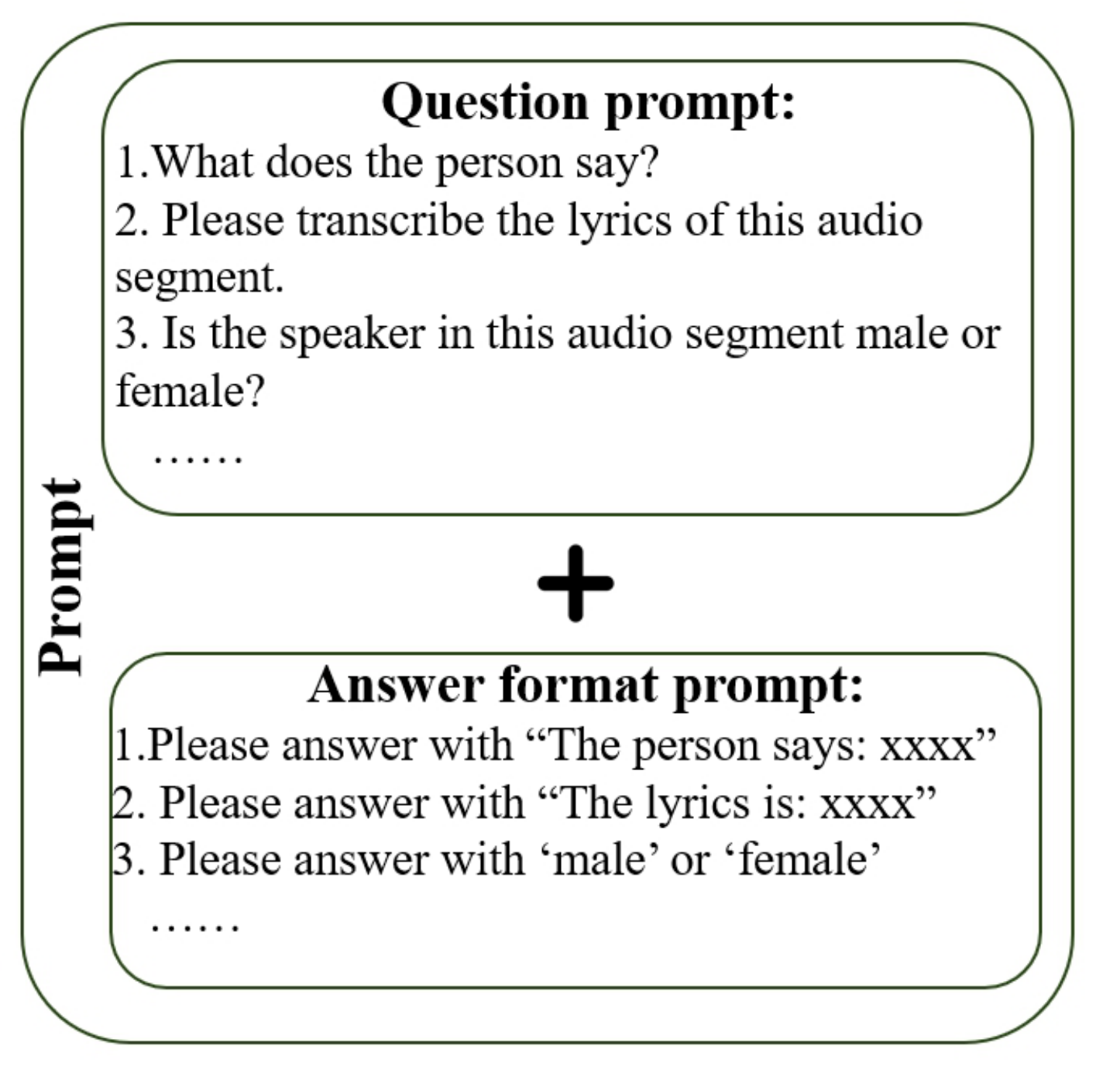}
    \caption{The method to generate text instructions for the problems.}
    \label{fig:prompt_generation}
\end{figure}

For objective multiple-choice questions, we guide large models by including multiple-choice options within the questions to facilitate the generation of final results. For subjective response questions, we specified the main aspects around which the questions revolve and set suggested answers, although these do not require the model to produce results that are exactly identical, illustrated in Fig.~\ref{fig:prompt_generation}.

\subsubsection{Uniform Sampling Rate}
Considering the potential introduction of extraneous factors due to varying sampling rates of audio data, this paper standardizes all datasets to the one with the lowest sampling rate. Consequently, all test data is downsampled to 16,000 Hz.
\subsubsection{Uniform number of audio channels}
To standardize the format of the input audio, we converted all audio files for the tasks into mono channel, except for those in the Binaural Effect Perception sub-task.
\subsubsection{Uniform Audio Duration}
\label{duration-restricte}
Most speech LLMs \citep{qwen_audio,qwen2audio,mullama,tang2023salmonn} utilize the encoder from \cite{whisper}, which limits their maximum audio processing duration to 30 seconds. To ensure fairness, we have restricted the lengths of the audio inputs to a maximum of 30 seconds.
\subsubsection{Uniform Option Ratio}
For the multiclass classification problem, we performed data balancing. Taking binary classification tasks as an example, due to some limitations in the current models, they might always choose one option in binary classification tasks. If the data were unbalanced, such as 40\% for one option and 60\% for the other, different models that always pick the same option could yield very different results, even though their capabilities are similar. This is not what we want, so we balanced the data for all multiclass classification tasks. Please refer to Tab.~\ref{tab:Utterances} for detailed information.
\begin{table}[ht]
    \centering
    \caption{Utterances for Each Task}
    \begin{tabular}{@{}l p{7cm}@{}} 
        \toprule
        \textbf{Task} & \textbf{Utterances} \\
        \midrule
        Language Identification & German: 500, Spanish: 500, English: 500, French: 500, Italian: 500 \\
        Auto-Speech Recognition & English:2791\\
        ASR for  Legal Terms & Chinese:102\\
        ASR for Medical Terms & Chinese:203\\
        Auto-Lyrics Transcription & English: 868\\
        Volume Perception & Increasing: 512, Decreasing: 512 \\
        Pitch Perception &(80-150)Hz: 300, (180-250)Hz: 300\\ 
        Binaural Effect Perception & Left ear: 400, Right ear: 400 \\
        Ambient Noise Detection & Yes: 824, No: 824 \\
        Acoustic Scene Classification & Babble: 310, Copy Machine: 310, Neighbor: 310, Shutting Door: 315, Airport Announcements: 305, Munching: 300, Typing: 310, Air-Conditioner: 305, Vacuum Cleaner: 310 \\
        Speaker's Age & Teens to Twenties: 330, Thirties to Forties: 330, Fifties to Sixties: 330 \\
        Speaker's Gender & Female: 1410, Male: 1410 \\

        Speech Emotion Recognition & Happy: 200, Disgust: 200, Fearful: 200, Sad: 200, Surprised: 200, Angry: 200, Neutral: 100 \\
        Cappella Emotion Recognition & Angry: 184, Sad: 184, Happy: 184, Fearful: 184, Neutral: 92 \\
        Emotion Intensity Perception & Former: 143, Latter: 143 \\
        Emotion Translation & English: 325\\ 
        Singing Detection & Singing: 1012, Speech: 1012 \\
        COVID-19 Risk Detection& Yes:56, No:64\\
        Cough Type Classification& Wet: 300 , Dry: 300\\
        Cough Origin Diagnosis&  COVID-19: 198, Healthy Cough: 200, Lower Infection: 200,Upper Infection: 200\\
        Cough Severity Assessment& Pseudocough: 170, Mild: 170, Severe: 170\\
        Spoken English Coach & English: 1009\\
        Voice Detective& English: 2134\\
        \bottomrule
    \end{tabular}
    \label{tab:Utterances}
\end{table}

\subsection{Data and Tools Utilized}
\label{appendix:data_tool_use}
We used the following 10 datasets:

Europarl-ST~\citep{Europarl-ST} ,LibriSpeech~\citep{LibriSpeech},JamendoLyrics MultiLang dataset ~\citep{Jam-Lyrics}, LJSpeech~\citep{ljspeech},Noisy speech~\citep{noisySpeech},SpeechAccentArchive~\citep{SpeechAccentArchive} ,VCTK~\citep{vctk},RAVDESS(~\citep{ravdess},AISHELL-MDSC~\citep{aishellMDSC},speechocean762~\citep{speechocean762}

We utilized two open-source tools:
 MS-SNSD~\citep{MSSNSD},cosyVoice~\citep{funaudiollm}


\subsection{Data Structure of Benchmark}
\label{sec:benchmark_data_structure}
Data samples are represented as (P, Q, A, D), where P denotes the audio path, Q represents the question, A corresponds to the answer, and D provides additional explanations to aid researchers in understanding the data.

\subsection{Details of Each Task}
\label{appendix:task_detail}
\subsubsection{Language Identification}
We used Europarl-ST \citep{Europarl-ST} to construct our evaluation dataset. Europarl-ST is a multilingual speech translation corpus containing paired audio-text samples for speech translation. It was constructed using debates held in the European Parliament between 2008 and 2012. We selected five commonly used languages: German, English, French, Spanish, and Italian. The task was set as: ``What language is spoken in this audio segment? Please choose from the German, English, French, Spanish and Italian options."
\subsubsection{Automatic Speech Recognition}
We constructed our evaluation dataset based on LibriSpeech \citep{LibriSpeech}. Inspired by \cite{whisper}, we used the test-clean and test-other splits as our test sets, comprising a total of 2791 data entries. Since we addressed specific aspects within our metric~\ref{appendix:wer}, we did not perform any additional processing when constructing the dataset. The task was set as: ``What does the person say? Please answer with `The person says: xxxx'."

\subsubsection{ASR for Legal Terms}
We selected 27 offenses defined under Chinese criminal law and combined them with four templates to generate 108 sentences, which were synthesized using cosyVoice \citep{funaudiollm}. After manual screening (detailed in Sec.~\ref{app:cred_verify_law}), 102 utterances remained. The task was set as: ``What does the person say? Please answer with `The person says: xxxx'." This approach is consistent with ASR, as we believe that this ability should be demonstrated automatically during the ASR process without the need for additional prompts.

\subsubsection{ASR for Medical Terms}
We selected 62 medical terms referring to specific locations and combined them with four templates to generate 248 sentences, which were synthesized using cosyVoice \citep{funaudiollm}. After manual screening (detailed in Sec.~\ref{app:cred_verify_law}), 203 utterances remained. The task was set as: ``What does the person say? Please answer with `The person says: xxxx'." This approach is consistent with ASR, as we believe that this ability should be demonstrated automatically during the ASR process without the need for additional prompts.
\subsubsection{Automatic Lyrics Transcription}
We utilized the JamendoLyrics MultiLang dataset \citep{Jam-Lyrics} for our research. We acknowledge that a revised version of this dataset has been released as the Jam-Alt dataset ~\citep{jam-alt}. However, in accordance with the constraints outlined in Sec.~\ref{duration-restricte}, we were required to resegment the audio files. Given that the Jam-Alt dataset, as described by its authors, exhibits certain deviations in its timestamps, we elected to employ the JamendoLyrics MultiLang dataset as our primary dataset for construction purposes. During the construction process, we manually selected the segmentation points and employed code to segment the audio files, thereby obtaining our final dataset.The task was set as: ``Please transcribe the lyrics of this audio segment.Please answer with: `The lyrics is: xxxx'."

\subsubsection{Volume Perception}
\label{appendix:volume_perception}
We constructed our evaluation dataset based on LJSpeech~\citep{ljspeech}. Following the data split of \cite{mult-speaker}, we used 512 test samples. We set up two scenarios: one where the volume gradually increases from 0 to its original level, and another where it decreases from the original level to 0. We tasked the model with determining whether the volume is increasing or decreasing. The task was set as: ``Is the volume of this audio segment gradually increasing or decreasing?"
\subsubsection{Pitch Perception}

We used the SpeechAccentArchive~\citep{SpeechAccentArchive} dataset to construct our test set. During this process, we first identified the frequency ranges with the highest proportion of fundamental frequency (F0). Ultimately, we selected the ranges (80, 150) Hz and (180, 250) Hz for our experiments. We framed the problem as follows: ``In the following audio segment, into which range does more than 70\% of the fundamental frequency content fall? Please choose from the following two ranges: (80, 150) Hz and (180, 250) Hz." We calculated the proportion of F0 content falling within these two ranges for each audio segment and selected the corresponding data. During the process, we ranked all the data, prioritizing those segments with a higher proportion.

\subsubsection{Binaural Effect Perception}
\label{appendix:binaural_Effect}
We generated random sounds using four methods: sine wave, square wave, triangle wave, and noise. These sounds are heard only in the left ear or the right ear. For more details, please refer to our public code. The model is used to determine which ear hears these sounds. The task was set as: ``In this audio segment, does the sound appear in the left ear or the right ear? Please answer with `left' or `right'."

\subsubsection{Ambient Noise Detection}
We constructed the evaluation dataset using Noisy speech~\citep{noisySpeech}.Noisy speech dataset contains corresponding pairs of noisy and clean data. The purpose of the dataset is to explore methods for speech enhancement.We selected the entire test set from this dataset, which includes 824 clean audio clips and 824 audio clips with ambient noise. We used all of these data, and the task was set as: ``Is there any ambient noise in this audio segment, in addition to the speaker voice? Please answer with yes or no."

\subsubsection{Acoustic Scene Classification}
We used  MS-SNSD~\citep{MSSNSD} to synthesize these test datasets.MS-SNSD is a tool for synthesizing speech with environmental noise, aimed at advancing research in speech enhancement. We selected 51 environmental noise samples from its test set to synthesize 6,105 test samples, and the task was set as: ``What is the ambient noise of this audio segment? Please choose from the ['Babble', 'CopyMachine', 'Neighbor', 'ShuttingDoor', 'AirportAnnouncements', 'Munching', 'Typing', 'AirConditioner', 'VacuumCleaner'] options?"

\subsubsection{Speaker’s Age Prediction}
We have observed that there are relatively few datasets specifically aimed at speaker age recognition. We noted that the AIR Bench~\citep{airbench} has done an excellent job in addressing this task,We followed their approach of categorizing age into four groups but noticed that their data distribution was not balanced, specifically: teens to twenties: 653, thirties to forties: 268, fifties to sixties: 64, seventies to eighties: 15. Therefore, we used the SpeechAccentArchive~\citep{SpeechAccentArchive} to balance the age distribution. Unfortunately, we found it difficult to obtain sufficient data for the seventies to eighties category, so we retained only three categories: teens to twenties, thirties to forties, and fifties to sixties. And the task was set as: ``Which age range do you believe best matches the speaker's voice? Please choose from the [`teens to twenties', `thirties to forties', `fifties to sixties'] options?"

\subsubsection{Speaker’s Gender Recognition}
We constructed the evaluation dataset using VCTK~\citep{vctk}.To balance the number of males and females in the benchmark, considering there are 61 female speakers and 47 male speakers in the VCTK dataset, we selected the top 47 female speakers along with all the male speakers. For each speaker, we chose the first 30 audio recordings. The task was set as: ``Is the speaker in this audio segment male or female?Please answer with `male' or `female'."

\subsubsection{Speech Emotion Recognition}
In a genuine sense, understanding emotions in models should not solely depend on interpreting text. Emotions do not have a one-to-one correspondence with sentences; the same sentence can express various emotional tones depending on the speaker's emotional state. Therefore, it is crucial to advocate for models to move beyond mere textual content of sentences when inferring emotions and to delve into the non-textual information within the speech. Accordingly, in the evaluation set for emotion recognition, we employed a dataset unrelated to both the emotions and the sentence content—the RAVDESS dataset ~\citep{ravdess}. The task is then defined as: ``What emotion does this audio clip convey? Please answer by single word select from [`neutral', `happy', `sad', `angry', `fearful', `disgust', `surprised']."

To demonstrate that the emotions in our constructed dataset are independent of the textual content, we used a combination of the whisper-v3-large \citep{whisper} model and the gpt-4-o \citep{OpenAI2023GPT4} model to predict the emotions in the audio files of the dataset. The experimental results can be found in the Tab.~\ref{tab:emotion-detection-Supplementary}

\begin{table}[htbp]
\centering
    \caption{emotion detection evaluation set Supplementary experiments}
\begin{tabular}{cccc}
\hline
 & \textbf{First repetition} & \textbf{Second repetition} & \textbf{Third repetition}\\

Accuracy & 10.53\% & 9.33\% & 9.73\% \\ \hline
\end{tabular}

\label{tab:emotion-detection-Supplementary}
\end{table}

\subsubsection{Cappella Emotion Recognition}
\label{appendix_singing_recognition}
We also used RAVDESS \citep{ravdess} to construct the evaluation set for singing emotion detection.The task is then defined as: ``What emotion does this audio clip convey? Please answer by single word select from [`neutral', `happy', `sad', `angry', `fearful', `disgust', `surprised']."

\subsubsection{Emotional Intensity Perception}
\label{appendix_emo_stength}
We used the RAVDESS \citep{ravdess} dataset to construct the evaluation set for Emotional Intensity Perception. Since most models accept only a single audio input, we merged two audio segments and tasked the model with analyzing which part of the combined audio segment exhibits stronger emotional intensity. Specifically, we defined the problem as follows: ``In this audio segment, a sentence is repeated twice. Is the emotion in the `former' stronger or the `latter' stronger? Please answer with `former' or `latter'." To balance the proportion between the two options, we alternated the placement of the stronger emotion, sometimes positioning it at the former and other times at the latter when synthesizing the data.

\subsubsection{Emotion Translation}
\label{appendix_emoT}
We believe that translations should reflect different expressions based on the emotional context. For example, the phrase ``What are you doing?" can convey various meanings depending on the emotion—whether it's anger, surprise, sadness, or neutrality. In an angry context, it expresses strong disapproval or questioning of the person's actions; in a surprised context, it conveys disbelief about what the other person is doing; and in a sad context, it should reflect disappointment. Therefore, translations should be adjusted accordingly to better capture these nuances.

We observed that cosyVoice ~\citep{funaudiollm} demonstrates excellent zero-shot capabilities, effectively mimicking the tone and style of the input speech prompt. Therefore, we used cosyVoice to emulate the sentences with strong emotions from the RAVDESS ~\citep{ravdess} dataset to generate speech with corresponding emotions. After synthesis, we had five native speakers review the generated speech. If any of the native speakers felt that the synthesized speech did not convey the intended emotion, that segment was discarded. Ultimately, we obtained xxx valid speech samples. The task was set as: ``Please translate the following sentence into the most appropriate Chinese, based on the emotion and content of this audio segment."
\subsubsection{Singing Detection}
We aim for singing detection to go beyond simply identifying background music or relying on lyrics to determine whether singing is occurring. Instead, we seek to differentiate singing from normal speech by recognizing the distinct rhythm and melody of singing. To achieve this, we constructed our singing detection dataset using RAVDESS (~\citep{ravdess}), which consists entirely of a cappella performances where the context is unrelated to the singing. The task is then defined as: ``Is there singing in this audio clip?Please answer by yes or no.".

\subsubsection{COVID-19 Risk Detection}
\label{appendix_covid19}
We use the Virufy COVID-19 Open Cough Dataset \citep{virufy} to construct our evaluation set. We classify the samples with positive test results as COVID-19 at risk, while those with negative results are classified as not at risk. And the task was set as: ``Please listen to the following cough sound and determine whether the person is at risk of having a COVID-19 infection. Respond with yes or no."

\subsubsection{Cough Type Classification}
We use the COUGHVID \citep{coughvid} dataset to construct our evaluation set. We only utilize the data that has been assessed by experts, which falls into two categories: evaluations by four experts and evaluations by one expert. We prioritize samples where three out of four experts agree, and then we use samples rated as "good" by the single expert. In this task, we ask the model to distinguish whether the cough is a wet cough or a dry cough. And the task was set as: ``Please help me determine whether the cough in this audio segment is a dry cough or a wet cough. Please respond with `wet' or `dry'."

\subsubsection{Cough Origin Diagnosis}
We use the COUGHVID \citep{coughvid} dataset to construct our evaluation set. We only utilize the data that has been assessed by experts, which falls into two categories: evaluations by four experts and evaluations by one expert. We prioritize samples where three out of four experts agree, and then we use samples rated as "good" by the single expert. In this task,The origins we tested include`COVID-19', `healthy cough', `lower infection', or `upper infection'. And the task was set as: ``Please help me determine the infection origin of the cough in the following audio segment. Choose from `COVID-19', `healthy cough', `lower infection', or `upper infection'."

\subsubsection{Cough Severity Assessment}
We use the COUGHVID \citep{coughvid} dataset to construct our evaluation set. We only utilize the data that has been assessed by experts, which falls into two categories: evaluations by four experts and evaluations by one expert. We prioritize samples where three out of four experts agree, and then we use samples rated as "good" by the single expert. In this task,
the severity levels we tested include: `pseudocough', `mild', or `severe'. And the task was set as: ``Please help me assess the severity of the cough in the audio segment. Choose from `pseudocough', `mild', or `severe'."

\subsubsection{Spoken English Coach}
\label{appendix_spoken_coach}
We used speechocean762~\citep{speechocean762} to construct our evaluation set.In selecting our evaluation set, we aimed to include a wide variety of pronunciation errors by prioritizing sentences with poorer pronunciation quality. Here is how we built our sentence collection:
We started by selecting 207 sentences based on word stress errors (score == 5).
Next, we chose 6 sentences with incomplete sentences or error-containing words (score \textless{} 10).
Then, we selected 332 sentences with poor fluency (score \textless{}= 5).
Following that, we picked 85 sentences with poor rhythm (score \textless{}= 5).
Subsequently, we chose 179 sentences with low accuracy (score \textless{}= 5).
Finally, we selected 40 sentences from each accuracy score level where the scores were higher.
This process resulted in a final set of 1009 sentences.
When constructing the ground truth for the answer output, we adopted the descriptions used in the original project for dataset scoring, and by concatenating these descriptions, we formed the final answer.
\subsubsection{Voice Detective}
\label{voice_detective}
When constructing the Voice Detective evaluation set, we used the SpeechAccentArchive dataset ~\citep{SpeechAccentArchive}. The primary reason for choosing this dataset is the difficulty in obtaining a large amount of similar data, which significantly reduces the risk of data leakage. This constraint also compels researchers to focus more on factors such as the age and background of the users within the dataset.

\subsection{Credibility Verification}
\label{app:cred_verify}
\subsubsection{ASR for Legal Term}
\label{app:cred_verify_law}
Since the legal vocabulary we selected, can be found in open-source code, is not complex, we introduced only one evaluator with a background in legal education, who is a native Mandarin speaker. The remaining three evaluators are regular native Mandarin speakers, making a total of four evaluators. If any one of the evaluators deems the speech quality insufficient, the corresponding speech will be discarded. The specific details of the evaluators are as follows:

Evaluator 1: 24 years old, male, graduated with a bachelor's degree from China University of Political Science and Law and is currently a master student at China University of Political Science and Law. Native Mandarin speaker.

Evaluator 2: 20 years old, female, currently an undergraduate student at Hubei University of Technology. Native Mandarin speaker.

Evaluator 3: 20 years old, female, currently an undergraduate student at Wuchang Shouyi University. Native Mandarin speaker.

Evaluator 4: 26 years old, male, high school graduate. Native Mandarin speaker.

\subsubsection{ASR for Legal Medical}
Due to the involvement of some medical terminology, this paper selected two evaluators with a medical background, along with two additional evaluators without a medical background. All of them are native Mandarin speakers. Similarly, if any one of the evaluators finds an abnormality in the speech, it will be discarded. The specific details of the evaluators are as follows:

Evaluator 1: 33 years old, female, graduated with a bachelor's degree from Hebei Medical University and has since been working in a medical-related field. Native Mandarin speaker.

Evaluator 2: 26 years old, female, completed an eight-year integrated program (continuously pursued both bachelor's and master's degrees) at Hebei Medical University and continues to work in a medical-related field. Native Mandarin speaker.

Evaluator 3: 25 years old, male, graduated with a bachelor's degree from Beijing Forestry University and is currently a graduate student at Beijing University of Posts and Telecommunications. Native Mandarin speaker.

Evaluator 4: 54 years old, male, graduated from a technical secondary school. Native Mandarin speaker.

\subsubsection{Emotion Translation}
We selected four evaluators and recorded their English proficiency. Similarly, if any one of the evaluators finds an abnormality in the speech, it will be discarded. The specific details of the evaluators are as follows:

Evaluator 1: 25 years old, female, graduated with a bachelor's degree from China Jiliang University and a master's degree from Beijing University of Posts and Telecommunications. English proficiency: CET-6.

Evaluator 2: 25 years old, female, graduated with both a bachelor's and a master's degree from Beijing University of Posts and Telecommunications. English proficiency: CET-6.

Evaluator 3: 23 years old, male, graduated with a bachelor's degree from Beijing Institute of Technology and is currently a PhD student at The Chinese University of Hong Kong, Shenzhen. English proficiency: IELTS Academic score: 6.5.

Evaluator 4: 28 years old, male, graduated with a bachelor's degree from Beijing University of Posts and Telecommunications and is a PhD student at Beijing University of Posts and Telecommunications. English proficiency: CET-6.

\section{Experiment Details}
\label{sec:Experiment_Details}
Below, we will divide the experiment details into four parts: details of human evaluation in Sec.~\ref{sec:humans_detail}, details of model evaluation in Sec.~\ref{appendix:model-evluation}, and metric details in Sec.~\ref{appendix:matric}.
\subsection{Humans Evaluation Details}
\label{sec:humans_detail}
In this section, we will introduce the participant information of our humans performance evaluation in Sec.~\ref{sec:part_inf} and present the results of the consistency test for the result in Sec.~\ref{sec:consistency_test}.
\subsubsection{Participant Information}
\label{sec:part_inf}
Evaluator 1: Female, 28 years old, graduated with a bachelor's degree from East China Normal University, PhD from the Institute of Physics CAS.

Evaluator 2: Female, 26 years old, graduated with a bachelor's degree from Beijing Normal University, master's degree from Shanghai Jiao Tong University.

Evaluator 3: Male, 29 years old, graduated with a bachelor's degree from Beijing University of Chemical Technology, PhD from Beijing University of Posts and Telecommunications.

Evaluator 4: Male, 27 years old, graduated with a bachelor's degree from Xidian University, currently pursuing a PhD at Singapore University of Technology and Design.

\subsubsection{Consistency Test}
\label{sec:consistency_test}
To verify the consistency of the humans evaluation, We focus on objective multiple-choice questions. we calculated the proportion of questions where all three volunteers selected the same option, as well as the proportion where all four volunteers chose the same option, relative to the total number of questions. These proportions are shown in Tab.~\ref{human_test_consistency}.

\begin{table}[h]
\small
\caption{Consistency for Humans Evaluation}
\label{human_test_consistency}
\begin{tabular}{@{}lcccc@{}}
\toprule
\multirow{2}{*}{\textbf{Task}}    & \multirow{2}{*}{\textbf{Accuracy}}  & \multirow{2}{*}{\textbf{Num of Questions}} & \textbf{Proportion}  & \textbf{Proportion} \\ 
&  & & (3 Evaluators Same) &  (4 Evaluators Same) \\ 
\midrule
Volume Perception              & 100.00\% & 160                        & 100.00\%                              & 100.00\%                              \\
Pitch Perception               & 96.25\%  & 160                        & 100.00\%                              & 95.00\%                               \\
Binaural Effect Perception     & 100.00\% & 160                        & 100.00\%                              & 100.00\%                              \\ 
Ambient Noise Detection        & 91.88\%  & 160                        & 100.00\%                              & 87.50\%                               \\
Acoustic Scene Classification & 90.28\%  & 720                       & 97.22\%                               & 93.89\%                               \\
Speaker’s Age Prediction       & 52.59\%  & 240                        & 76.67\%                               & 46.67\%                               \\
Speaker’s Gender Recognition   & 97.50\%  & 160                        & 100.00\%                              & 100.00\%                              \\
Speech Emotion Recognition     & 50.71\%  & 560                       & 94.29\%                               & 85.71\%                               \\
Cappella Emotion Recognition   & 62.25\%  & 400                       & 92.00\%                               & 68.00\%                               \\
Emotion Intensity Perception   & 97.50\%  & 160                        & 100.00\%                               & 95.00\%                              \\
Singing Detection              & 98.13\%  & 160                        & 100.00\%                              & 97.50\%                               \\
COVID-19 Risk Detection        & 60.63\%  & 160                        & 70.00\%                               & 17.50\%                               \\
Cough Type Classification      & 52.50\%  & 160                        & 77.50\%                               & 22.50\%                               \\
Cough Origin Diagnosis         & 32.19\%  &   320                      & 28.75\%                                & 2.50\%                               \\
Cough Severity Assessment      & 45.42\%  & 240                        & 45.00\%                               & 11.67\%                               \\ \bottomrule
\end{tabular}
\end{table}

It is also important to note that, since our testers are only proficient in English, they were unable to complete the Language Identification task.

\subsubsection{Deficiency in Humans Evaluation.}
During the Humans Evaluation process, we were unable to find a native English speaker, but all participants involved in the evaluation are proficient English users.
We also could not find individuals who are proficient in multiple languages, which made it difficult to conduct a Humans Evaluation for the Language Identification task.

\subsection{Models Evaluation Details}
\label{appendix:model-evluation}
We divide our experimental details into two sections: the model replication platform in Sec.~\ref{appdenix:model-platform}, and the model replication details in Sec.~\ref{appendix:model-replication}.

\subsubsection{Experimental Platform}
\label{appdenix:model-platform}
In this paper’s experiments, all servers used are equipped with an Intel® Xeon® Platinum 8358 CPU @ 2.60GHz as the core processor. Each server is loaded with eight NVIDIA A800-SXM4-80GB graphics cards, and each model runs with exclusive use of one A800 card.

\subsubsection{Models Replication Details}
\label{appendix:model-replication}
In this paper, we aim to select the 7B-level versions of various models wherever possible. However, due to the differences between various models, it is difficult to ensure that their parameter counts are exactly the same.

\textbf{GPT-4o}
For the GPT-4o model, we reproduced the model by calling its API.

\textbf{Mu-LLaMA}
In the process of implementing the model Mu-LLaMA~\citep{mullama} , this paper used the LLama2-7B-chat~\citep{llama2} checkpoint to maintain consistency with the original paper, and utilized the open-source MU-LLaMA checkpoint provided.

\textbf{GAMA}
Since the primary focus of this paper is to test the audio understanding capabilities of the GAMA model \citep{GAMA}, we consulted with the authors and selected the `state4epoch2' checkpoint over the `state5epoch2' checkpoint, as it has superior audio comprehension abilities

\textbf{SALMONN}
For the SALMONN model \citep{tang2023salmonn}, we tested the model using its open-source code.

\textbf{Qwen2-Audio}
For the Qwen2-Audio model \citep{qwen2audio}, we reproduced the model using the 7B version of its open-source code.

\subsection{Matrix}
\label{appendix:matric}
We have designed three metrics: WER, the accuracy for objective multiple-choice questions, and GPT-4o scoring, specifically targeting ASR tasks, objective multiple-choice questions, and subjective responses. This section will provide detailed explanations. For an overview, please refer to the following Tab.~\ref{tab:Metrics}.

\begin{table}[ht]
    \centering
    \caption{Metrics for Each task}
    \begin{tabular}{@{}l c@{}} 
        \toprule
        \textbf{Task} & \textbf{Metric} \\
        \midrule
        Language Identification & 5-Categories Acc \\
        Speech ASR & WER \\
        Song ASR & WER \\
        Volume Perception & 2-Categories Acc \\
        Binaural Effect Perception & 2-Categories Acc \\
        Ambient Noise Detection & 2-Categories Acc \\
        Speaker's Age & 3-Categories Acc \\
        Speaker's Gender & 2-Categories Acc \\
        Sound Event Classification & 9-Categories Acc \\
        Singing Detection & 2-Categories Acc \\
        Speech Emotion Recognition & 7-Categories Acc \\
        Song Emotion Recognition & 5-Categories Acc \\
        Emotion Intensity Perception & 2-Categories Acc \\
        Disorder Detection & 2-Categories Acc \\
        Speech Disorders Detection & 2-Categories ACC \\
        COVID-19 Risk Detection & 2-Categories ACC \\
        ALS Detection & 2-Categories ACC \\
        Accent Detection & 11-Categories Acc \\
        Emotion Translation & GPT Score \\
        Spoken English Coach & GPT Score \\
        Voice Detective & GPT Score \\
        \bottomrule
    \end{tabular}
    \label{tab:Metrics}
\end{table}

\subsubsection{WER for ASR}
\label{appendix:wer}

The Word Error Rate (WER), a key metric  for gauging the effectiveness of Automatic Speech Recognition (ASR) systems, quantifies the divergence between an ASR system's output and a reference transcript. It assesses the total error rate by tallying the number of insertion, deletion, and substitution operations needed to align the ASR output with the true reference text.

While computing the WER, certain variances in word usage, like "I am" compared to "I’m," may be seen as semantically equivalent by human standards but are flagged as errors by computational algorithms. Thus, a standardization process is essential prior to WER calculation to make both texts directly comparable. The methodology for this standardization, akin to what is employed in the Whisper~\citep{whisper} framework, has been detailed in a related research paper. It has been demonstrated that this approach exerts negligible influence on the assessment of WER outcomes when tested against the LibriSpeech~\citep{LibriSpeech} dataset, which was utilized in our paper.

For cases where the error rate exceeds 100\% (i.e., WER is over 1), we mark them in our experimental records as having significant recognition errors. Such data will not be included in the calculation of the final average WER. In the final record of the experiment, we will focus on two key metrics: first, the ASR completion rate, which is the percentage of data with a WER less than 1; second, the mean WER of the completed portion, which is the average WER of data with a WER less than 1. If the mean WER of the completed portion does not decrease to below 0.8, we will conclude that the model lacks effective automatic speech recognition (ASR) capabilities and document this finding in detail in the experimental results.

The implementation details regarding WER (Word Error Rate) can be found in our publicly available code.

\subsubsection{Accuracy for objective multiple-choice questions}
A selection is considered correct only if the model chooses the correct answer and no other options. If the model selects two or more options, even if the correct one is included, it will be deemed incorrect.

\subsubsection{Accuracy for ASR on Terms}

Since in these tasks we primarily assess the ability of speech LLMs to transcribe terms, we consider a response correct as long as the correct term is included in the speech transcription, without focusing on the accuracy of other parts of the sentence.

\subsubsection{Scoring for Subjective Response Questions}
\label{appendix:gpt-4o}
In our experiments, we used GPT-4o to assist in evaluating the results. The specific prompt used is as follows.

\textbf{Prompt for Emotion Translation}

    I currently need your assistance in evaluating some translations. The most suitable translations should incorporate the corresponding emotions appropriately. The scoring ranges from 0 to 4. I will provide you with the original English sentence, the associated emotional label, and the suggested translation, allowing you to score them based on the context.
    
    Here are some examples:

    [Here are some scoring examples. Due to space limitations, we have omitted them in this section. You can find the details in the code we have made available.]
    
    Now Answer:[ANSWER]
    
    Label:The original sentence is: \textless emotion\textgreater [SENTENCE] The suggested translation is:  [SUGGESTION]. 
    
    Please provide your score.

\textbf{Prompt for Spoken English Coach}

    I now need you to help me evaluate some Answers for accuracy.
    You need to evaluate and score in the order of overall pronunciation, fluency, prosody, words that are mispronounced, and words that have incorrect stress. The score ranges from 0 to 4. Here are the specific scoring rules:
    You need to first check if the evaluation of overall pronunciation in the Answer matches the Label. If they do not match, give a score of 0 and continue with the evaluation; if there is no relevant description, also give a score of 0 and continue with the evaluation; if it is correct, add 1 point and continue with the evaluation.
    
    For fluency and prosody in the Answer compared to the Label, award up to 1 point for each if completely correct, a partial score for partially correct, and no points if there is no relevant expression.
    Finally, check the descriptions in the Answer and Label regarding words that are mispronounced and words that have incorrect stress. Award 1 point only if all are correct. If part of the descriptions are correct, you can give a partial score, such as 0.33 points for one out of three correct descriptions.
    Here are some examples:

    [Here are some scoring examples. Due to space limitations, we have omitted them in this section. You can find the details in the code we have made available.]
    
    Now Answer:[ANSWER]
    
    Label:[LABEL]
    
    Please provide your score.

\subsubsection{Prompt for Voice Detective}

    I now need you to help me evaluate some Answers for accuracy.
    You should focus on whether the information about gender, place of birth, age, and native language in the Answer matches the Label, and provide a final rating.
    Award 1 point for each correct piece of information, with no points for incorrect information. Please give your score on a scale of 0 to 4.
    Here are some examples:

    [Here are some scoring examples. Due to space limitations, we have omitted them in this section. You can find the details in the code we have made available.]
    
    Now Answer:[ANSWER]
    
    Label:[LABEL]
    
    Please provide your score.

\subsection{Speech Instruction}
\label{app:speech-instruction}

When adopting the speech instruction, we use Google Translate's text-to-speech tool to convert the text instruction into speech, which is then merged with the original audio segment and fed into the speech LLMs.

\section{Instruction Follow Experiment}
\label{appendix:prompt}

\subsection{Speaker’s Age Prediction}
The instructions used in the experiment are as follows:

\begin{itemize}
\item \textbf{Instruction variation I} In which age group do you think the speaker's voice belongs?
\item \textbf{Instruction variation II} What age category do you believe the speaker's voice fits into best?
\item \textbf{Instruction variation III} Which age bracket do you feel corresponds to the speaker's voice?
\item \textbf{Instruction variation IV} How old do you think the speaker sounds, based on their voice?
\item \textbf{Instruction variation V} Which age range would you assign to the speaker's voice?
\item \textbf{Instruction variation VI} What age range do you associate with the speaker’s voice?
\item \textbf{Instruction variation VII} Which age group do you think best describes the speaker’s vocal characteristics?
\item \textbf{Instruction variation VIII} What do you believe is the age range of the speaker judging by their voice?
\end{itemize}
The experimental results are recorded in Tab.~\ref{tab:prompt_age}.

\begin{table}[h]
\small
\centering
    \caption{The impact of different prompts on age detection}
    \label{tab:prompt_age}
\begin{tabular}{lcccc}
\hline
 \textbf{Prompt}& \textbf{Qwen-Audio} & \textbf{Qwen2-Audio} & \textbf{MuLLama}& \textbf{GAMA}\\
 \hline
Our benchmark instruction & 29.29\% & 38.55\% & 33.60\% & 0.2\%\\ 
Instruction variation I & 23.03\% & 36.36\% & 35.45\%& 0.4\% \\
Instruction variation II & 31.82\% & 36.97\% & 35.45\%& 4.85\% \\ 
Instruction variation III & 12.83\% & 38.38\% & 34.75\%& 0.0\% \\
Instruction variation IV & 4.44\% & 43.03\% & 31.31\%& 0.2\% \\ 
Instruction variation V & 28.89\% & 37.37\% & 33.03\%& 0.1\% \\ 
Instruction variation VI & 19.90\% & 37.27\% & 34.14\%& 0.0\% \\ 
Instruction variation VII & 6.57\% & 36.77\% & 30.81\%& 0.3\% \\ 
Instruction variation VIII & 26.77\% & 41.11\% & 28.67\%& 0.4\% \\ \hline

\end{tabular}
\end{table}

\subsection{Acoustic Scene Classification}

\begin{itemize}
\item \textbf{Instruction variation I} How would you detect the background sound in this audio clip?
\item \textbf{Instruction variation II} What kind of ambient noise can be heard in this segment?
\item \textbf{Instruction variation III} Can you describe the environmental sounds present in this audio?
\item \textbf{Instruction variation IV} What background audio elements are featured in this segment?
\item \textbf{Instruction variation V} What atmosphere is created by the sounds in this audio segment?
\item \textbf{Instruction variation VI} Can you identify the ambient sound in this clip?
\item \textbf{Instruction variation VII} What noises are occurring in the background of this audio?
\item \textbf{Instruction variation VIII} What type of surrounding sound is present in this recording?
\end{itemize}
The experimental results are recorded in Tab.~\ref{tab:prompt_env}.

\begin{table}[h]
\small
\centering
    \caption{The impact of different prompts on acoustic scene classification}
    \label{tab:prompt_env}
\begin{tabular}{lcccc}
\hline
 \textbf{Prompt}& \textbf{Qwen-Audio} & \textbf{Qwen2-Audio} & \textbf{MuLLama}& \textbf{GAMA}\\
 \hline
Our benchmark instruction & 18.84\% & 27.67\% & 5.07\% & 12.05\%\\ 
Instruction variation I & 13.05\% & 35.68\% & 1.91\%& 0.00\% \\
Instruction variation II & 8.97\% & 13.73\% & 5.91\%& 0.36\% \\ 
Instruction variation III & 4.29\% & 9.66\% & 0.00\%& 0.94\% \\
Instruction variation IV & 5.43\% & 9.95\% & 0.00\%& 1.87\% \\ 
Instruction variation V & 13.95\% & 28.29\% & 1.87\%& 0.54\% \\ 
Instruction variation VI & 15.32\% & 21.87\% & 2.02\%& 0.25\% \\ 
Instruction variation VII & 5.37\% & 5.23\% & 1.8\%& 0.00\% \\ 
Instruction variation VIII & 9.62\% & 18.92\% & 6.31\%& 4.32\% \\ \hline

\end{tabular}
\end{table}

\end{document}